\documentclass[sigconf,nonacm]{acmart}
\usepackage{amsmath,amsfonts,bm}

\def\eqref#1{equation~\ref{#1}}
\def\1{\bm{1}}

\DeclareMathAlphabet{\mathsfit}{\encodingdefault}{\sfdefault}{m}{sl}
\SetMathAlphabet{\mathsfit}{bold}{\encodingdefault}{\sfdefault}{bx}{n}

\usepackage{booktabs}
\usepackage{enumitem}
\usepackage{hyperref}
\usepackage{multirow}
\usepackage{url}
\usepackage{amsmath}
\usepackage{graphicx}
\usepackage{subcaption} 
\usepackage{wrapfig}
\usepackage{dsfont}
\usepackage{array}
\usepackage{longtable}
\usepackage{stfloats}
\AtBeginDocument{%
  }

\begin{document}

\title[From Evidence to Trajectory: Abductive Reasoning Path Synthesis for RAG Agents Development]{From Evidence to Trajectory: Abductive Reasoning Path Synthesis for Retrieval-Augmented Generation Agents Development}

% \affiliation{%

\author{Muzhi Li}
\affiliation{%
  \institution{The Chinese University of Hong Kong}
  \city{Sha Tin}
  \state{NT}
  \country{Hong Kong}}
\email{mzli@cse.cuhk.edu.hk}
\orcid{0009-0008-1331-3061}

\author{Jinhu Qi}
\affiliation{%
  \institution{The Chinese University of Hong Kong}
  \city{Sha Tin}
  \state{NT}
  \country{Hong Kong}}
\email{jhqi25@cse.cuhk.edu.hk}
\orcid{0009-0006-5544-4786}

\author{Yihong Wu}
\affiliation{
  \institution{Université de Montréal}
  \city{Montréal}
  \state{QC}
  \country{Canada}
  }
\email{yihong.wu@umontreal.ca}
\orcid{0009-0009-2680-4107}

\author{Minghao Zhao}
\affiliation{%
  \institution{The Chinese University of Hong Kong}
  \city{Sha Tin}
  \state{NT}
  \country{Hong Kong}}
\email{mhzhao25@cse.cuhk.edu.hk}
\orcid{0000-0003-2871-0023}

\author{Liheng Ma}
\affiliation{
  \institution{McGill University \& Mila}
  \city{Montréal}
  \state{QC}
  \country{Canada}}
\email{liheng.ma@mail.mcgill.ca}
\orcid{0009-0005-8340-4813}

\author{Yifan Li}
\affiliation{%
  \institution{The Chinese University of Hong Kong}
  \city{Sha Tin}
  \state{NT}
  \country{Hong Kong}}
\email{yfli24@cse.cuhk.edu.hk}
\orcid{0009-0000-2414-367X}

\author{Xinyu Wang}
\affiliation{
  \institution{McGill University}
  \city{Montréal}
  \state{QC}
  \country{Canada}}
\email{xinyu.wang5@mail.mcgill.ca}
\orcid{0009-0004-0521-2077}

\author{Zhenghan Tai}
\affiliation{
  \institution{University of Toronto}
  \city{Toronto}
  \state{ON}
  \country{Canada}}
\email{winfred.tai@mail.utoronto.ca}
\orcid{0009-0003-3896-5131}

\author{Zixing Song}
\affiliation{
  \institution{University of Bristol}
  \city{Bristol}
  \state{}
  \country{United Kingdom}}
\email{zixing.song@bristol.ac.uk}
\orcid{0000-0002-8871-3990}

\author{Yingxue Zhang}
\affiliation{
  \institution{Huawei Noah’s Ark Lab}
  \city{Montréal}
  \state{QC}
  \country{Canada}}
\email{yingxue.zhang@huawei.com}
\orcid{0000-0002-8370-3873}

\author{Ho-fung Leung}
\affiliation{%
  \institution{Independent Researcher}
  \city{Sha Tin}
  \state{NT}
  \country{Hong Kong}}
\email{ho-fung.leung@outlook.com}
\orcid{0000-0003-4914-2934}

\author{Irwin King}
\affiliation{%
  \institution{The Chinese University of Hong Kong}
  \city{Sha Tin}
  \state{NT}
  \country{Hong Kong}}
\email{king@cse.cuhk.edu.hk}
\orcid{0000-0001-8106-6447}

\renewcommand{\shortauthors}{Muzhi Li et al.}

\begin{abstract}
  Retrieval-augmented generation (RAG) agent development is hindered by the lack of executable ground-truth agent-environment interaction trajectories. Existing datasets provide questions, answers, and evidence, but lack fine-grained supervision for retriever invocation, dynamic planning, and stepwise decision-making. Reinforcement learning offers a potential solution, but often suffers from sparse rewards and cold-start failures when base large language models~(LLMs) lack sufficient reasoning capability. Meanwhile, existing data synthesis methods mainly generate post-hoc rationales rather than executable environment-interaction trajectories. In this paper, we propose \textbf{EviPath}, an evidence-anchored reasoning path synthesis paradigm for RAG agent development. EviPath reverse-engineers executable trajectories from question-answer pairs and supporting evidence through three stages: (i) Abductive Subtask Planning, which decomposes questions and plans dependency-aware solution paths; (ii) Faithful Sub-question Answering, which uses supporting evidence as a proxy environment to generate grounded intermediate thoughts and answers; and (iii) Conversational Fine-Tuning, which converts complete trajectories into a dialogue-format for supervised fine-tuning. Experiments on widely used question-answering benchmarks show that an 8B model trained on our synthetic corpus significantly and consistently outperforms state-of-the-art baselines, achieving a \textbf{14.7\% absolute Exact Match gain} in open-domain question answering.
\end{abstract}




\maketitle

\section{Introduction}
Retrieval-augmented generation (RAG) agents, powered by large language models~(LLMs)~\citep{guo2025deepseek}, can autonomously decompose complex problems, gather external knowledge iteratively, and synthesize multi-hop answers. The capacity to operate with minimal human intervention enables RAG agents to adapt to various diverse applications, ranging from math problem solving~\citep{zhu-etal-2025-retrieval} and code generation~\citep{zhang-etal-2023-syntax} to financial analysis~\citep{wang2025finsagemultiaspectragfinancial,Veritasfi}. 

\begin{figure*}[t]
	\centering
	\includegraphics[width=\textwidth]{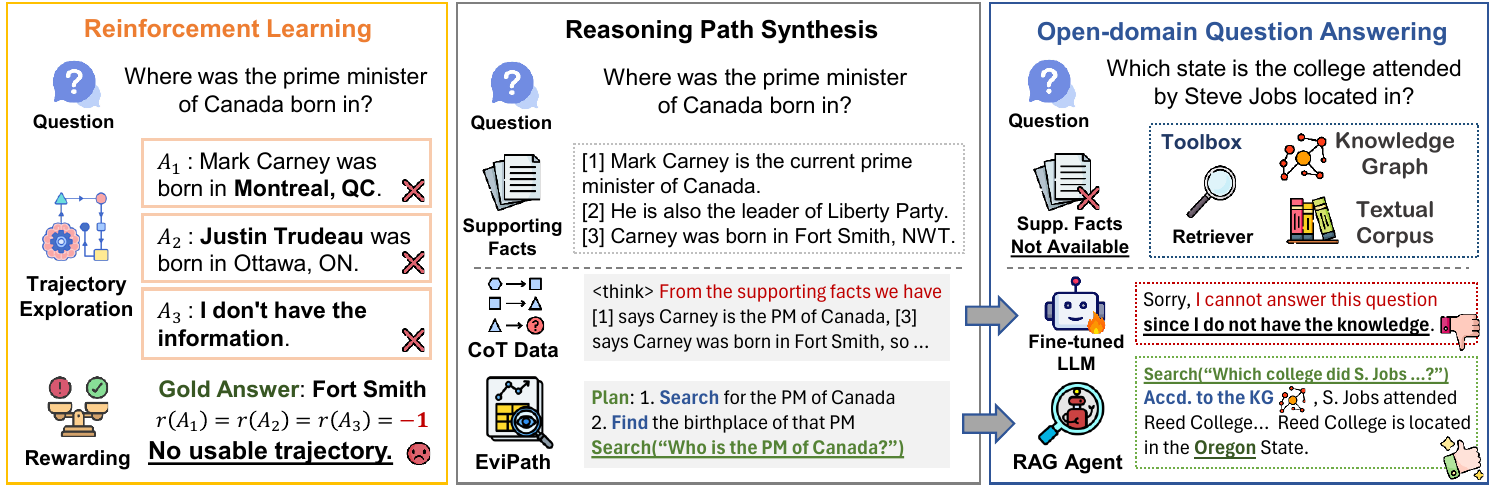}
    \caption{Limitations of prior paradigms for RAG agent development. Reinforcement learning-based methods often fail to discover usable trajectories when the base LLM lacks strong priors. CoT-style reasoning path synthesis offers static, post-hoc rationales over curated evidence and fails to model agent–environment interactions required in open-domain QA.}
	\label{fig:rl_synthesis}
\end{figure*}

Despite their promise, developing RAG agents remains challenging since\textbf{\textit{ ground-truth reasoning trajectories are unavailable}}. Mainstream multi-hop question answering datasets~\cite{HotpotQA,2Wiki} provide only final answers and supporting facts. They lack executable, step-wise supervision such as question decomposition, search query reformulation, and plan refinement.

One approach to training RAG agents without process supervision is reinforcement learning~(RL)~\citep{shao2024deepseekmath}, which allows LLMs to ``self-explore'' trajectories via trial and error. 
Nevertheless, these methods have notable limitations~(Figure~\ref{fig:rl_synthesis} left). 
First, the commonly-adopted outcome reward signals are often sparse and delayed, making it difficult to assign credit to individual actions.
In addition, online RL algorithms are inherently sample-inefficient and computationally expensive. To accurately estimate baselines for relative advantage calculation, they necessitate repeated sampling of multiple trajectories per query, wasting vast budgets on failed attempts merely to reduce variance~\citep{variance_reduction,shao2024deepseekmath}. 
Most critically, the effectiveness of RL relies heavily on the intrinsic reasoning capabilities of the model. This creates a severe ``cold-start'' problem for smaller models~(e.g., 1B-3B parameters). \textbf{\textit{Without sufficient prior knowledge, they typically fail to discover any correct actions that lead to a positive reward, rendering trajectory exploration ineffective}}. 

Another line of work~\citep{LongAlign,yang2025longfaith} mitigates data scarcity by synthesizing reasoning paths, leveraging LLMs to generate Chain-of-Thought~(CoT) rationales~(Figure~\ref{fig:rl_synthesis} mid and right). Yet, these approaches are ill-suited for agent training as they produce static, post-hoc explanations instead of dynamic and interactive decision-making processes. Consequently, \textbf{\textit{they fail to endow LLM with the core agentic capability to interact with external environments}}, limiting their effectiveness in \textbf{\textit{open-domain}} question answering.

To address the aforementioned limitations, we propose \textbf{EviPath}, an \textbf{Evi}dence-anchored reasoning \textbf{path} synthesis framework based on \textbf{\textit{abductive reasoning}}. Formally, abductive reasoning infers a plausible explanatory hypothesis from which the observations can be derived from the premise~\citep{bai-etal-2024-advancing}. In our setting, we view the question-answer pair as the \textit{observation} and the supporting evidence as the \textit{premise}, then ``reverse-engineer'' the reasoning trajectory without the need for blind exploration. Crucially, this approach bypasses the ``cold start'' problem by deriving the process from the result, rather than demanding the model to discover solutions from scratch. 
EviPath synthesizes these interactive trajectories via a three-stage pipeline following the Planner-Executor architecture~\citep{Search-o1}. Firstly, \textbf{\textit{(i) Abductive Subtask Planning}} uses supporting evidence to construct a dependency-aware reasoning plan, and then simulates iterative agent execution to generate planner-side thoughts and retrieval queries. 
Secondly, to mitigate the lack of agent--environment interaction in standard CoT trajectories synthesis frameworks, the \textbf{\textit{(ii) Faithful Sub-question Answering}} stage constructs a ``proxy environment'' with the provided supporting facts. It then grounds each sub-question with corresponding evidence, synthesizes reasoning thoughts and derives intermediate answers.
Finally, the \textbf{\textit{(iii) Conversational Fine-tuning}} stage packages the complete reasoning paths into a user-assistant dialogue format for supervised fine-tuning~(SFT). This offline, data-centric paradigm serves as a highly cost-effective alternative to RL, providing dense supervision without the need for expensive repeated sampling.
Extensive experiments show that an 8B-parameter agent trained on our synthetic corpus achieves \textbf{a double-digit absolute Exact Match gain of 14.7\%}, significantly outperforming state-of-the-art baselines. 
Our contributions can be summarized as follows:
\begin{itemize}[leftmargin=*]
    \item  We are the first to formulate the synthesis of reasoning paths for RAG agents as an \textbf{\textit{abductive reasoning problem}}. 
    This shifts the prevailing paradigm from ``\textit{forward trial-and-error}'' to ``\textit{reverse-engineering}'', providing a structured approach to generated reasoning trajectories without rely heavily on the intrinsic reasoning capabilities of base LLMs. 
    \item We propose \textbf{EviPath}, a \textbf{\textit{data-centric}} framework for RAG agent development. It solves the dual challenges of data scarcity and the reliance of expensive online RL, empowering smaller models (e.g., 1-8B) to acquire complex agentic capabilities via offline SFT.
    \item We construct \textbf{265k} golden reasoning trajectories. This scalable corpus is specifically designed to enhance agentic skills, such as high-level planning, retriever utilization, and context-aware reasoning for LLM post-training, serving as a high-quality asset for the community.
    \item We conduct extensive experiments on three widely used multi-hop QA datasets. The results show that RAG agents trained on EviPath-synthesized data significantly and consistently outperform all baselines, proving that synthetic data construction is a superior alternative to complex algorithmic engineering.
\end{itemize}

\section{Related Works}
\subsection{RAG Agents for QA}
RAG agents enhance LLMs with external evidence to mitigate hallucinations in knowledge-intensive QA~\citep{achiam2023gpt,cheng2024dated,hallucination}. The paradigm has evolved from simple ``retrieve-then-read'' pipelines~\citep{RAG,gao2023retrieval} to sophisticated workflows that interleave reasoning, tool use, and reflection~\citep{ReAct,IRCoT,SelfRAG,Iter-RetRAG}. Recent works scale agentic QA through learned monologues~\citep{IM-RAG}, and modular designs that separate planning from execution~\citep{Search-o1,Collab-RAG,RAG-Star}. Some efforts also leverage reinforcement learning to optimize policies of retrieval and reasoning~\citep{Search-R1,R1-Searcher,Mujica-MyGO}. Orthogonal advances strengthen the retrieval side, including query reformulation~\citep{RQ-RAG,RaFe}, end-to-end multi-hop retrieval~\citep{BeamRetrieval}, and knowledge graph integration~\citep{Graph-R1,KG-o1,DynaSearcher}. Despite progress, current methods rely heavily on outcome rewards or prompt heuristics. Consequently, when the base LLM lacks sufficient knowledge priors, it often fails to explore usable trajectories, rendering RL ineffective.

\subsection{Reasoning Path Synthesis}
Enhancing the reasoning capabilities of LLMs has garnered significant attention, driving the development of data synthesis methods~\citep{wang-etal-2023-self-prompted,xiong-etal-2024-effective,LongAlign,yang2025longfaith}. Earlier approaches like GENREAD~\citep{yu2023generateretrievelargelanguage} and SP-CoT~\citep{wang-etal-2023-self-prompted} focus on replacing retrieval with model-generated retrieval, but the synthesized reasoning paths are not grounded in evidence and thus remain vulnerable to hallucination. More recent works line in improving the long-context processing capabilities of LLMs by constructing continued pretraining data~\cite{xiong-etal-2024-effective,gao-etal-2025-train}, concatenating context segments into long training sequences to address the lost-in-the-middle problem~\citep{an2024make}, or generating step-wise CoT trajectories to answer complex, multi-hop questions~\citep{LongAlign}. In order to make the reasoning trajectories faithful and grounded, LongFaith~\citep{yang2025longfaith} proposes to include reasoning thoughts with chains of golden evidence citations, which effectively alleviates hallucinations and achieves desirable results. Nevertheless, these approaches focus on generating the chain-of-thought reasoning steps based on the fixed contexts, failing to guide the training of RAG agents that necessitate environment interactions.

\section{Problem Formulation}
\subsection{Answering Questions with RAG Agents}~\label{RAG_agent}
\begin{figure}[t]
    \centering
    \begin{minipage}{\linewidth}
        \centering
        \includegraphics[width=\linewidth]{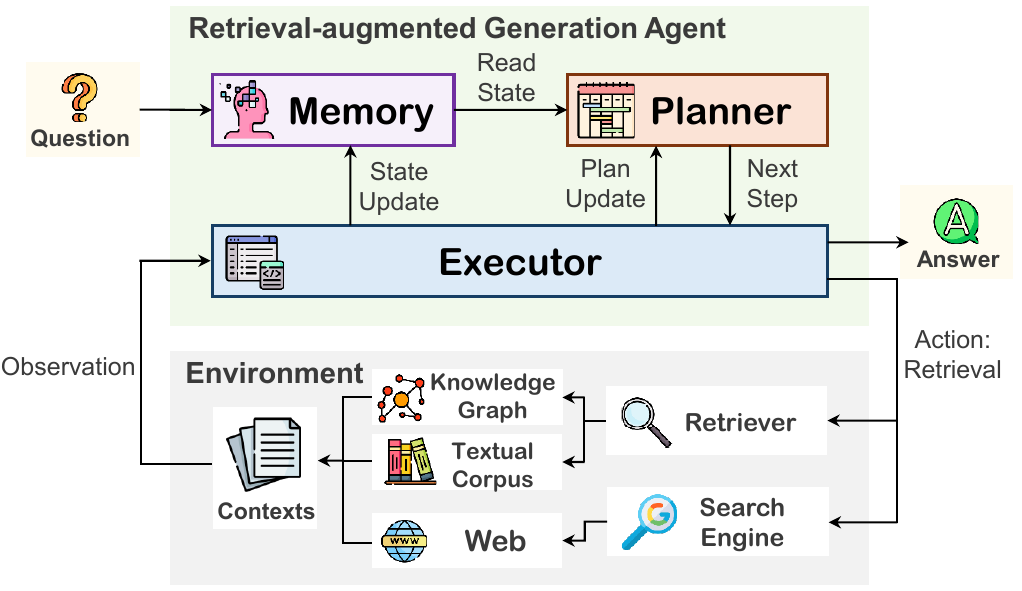}
        \caption{The planner-executor architecture of RAG agents.}
        \label{fig:rag_agent}
    \end{minipage}
\end{figure}
In this paper, we develop RAG agents to address the multi-hop question answering (MHQA) task in an open-domain setting. The core challenge of MHQA lies in aggregating evidence from diverse sources and conducting multi-step reasoning to derive the final answer. To address this, RAG agents (Figure~\ref{fig:rag_agent}) decouple the complex reasoning process into a hierarchical, two-level framework, which consists of a \textbf{\textit{Planner}} for high-level strategic planning and an \textbf{\textit{Executor}} for low-level sub-task execution.

The overall process is as follows: given a complex question $q$, the \textbf{\textit{Planner}} first formulates a plan $\mathcal{P}$ by decomposing $q$ into a sequence of atomic, solvable sub-questions. 
In each iteration $i$, the \textbf{\textit{Planner}} determines the specific set of sub-questions $Q_i$ to be resolved in the current step. 
Then, the \textbf{\textit{Executor}} resolves each sub-question $q_j\in Q_i$, interacts with an external knowledge base $\mathcal{K}$ to retrieve relevant context $\mathcal{C}_j$, and derives an answer $a_j$ for that sub-question. This process continues until all sub-questions are addressed. Finally, the \textbf{\textit{Planner}} synthesizes the intermediate results into a final answer $a$. Formally, the collaborative workflow can be expressed as:
\begin{align}
P(a|q,\mathcal{I},\mathcal{K})&=
\underbrace{P(\mathcal{P}|q, \mathcal{I})\cdot 
\prod_{i=1}^{|\mathcal{P}|} P\big(Q_i, \mathcal{R}_i \mid Q_{<i},a_{<i},q,\mathcal{P}\big)}_{\text{\small Planner}} \notag \\
&\cdot
\underbrace{\big(\!\prod_{q_j\in Q_i}\!P(\mathcal{C}_j|q_j,\mathcal{K}) \cdot 
P(a_j,\mathcal{R}_j^{(s)}|q_j,\mathcal{C}_j)\big),}_{\text{\small Executor}}
\label{eq:pf}
\end{align}

 where $\mathcal{I}$ denotes the instruction prompts, $|\mathcal{P}|$ denotes the number of reasoning steps, $a_{<i}$ contains the answer of all sub-questions prior to the $i$-th reasoning step; $\mathcal{R}_i$ denotes the reasoning thoughts made in $i$-th planning step, $\mathcal{R}_j^{(s)}$ denotes the thoughts for answering sub-question $q_j$.

\begin{figure*}[t]
	\centering
	\includegraphics[width=\textwidth]{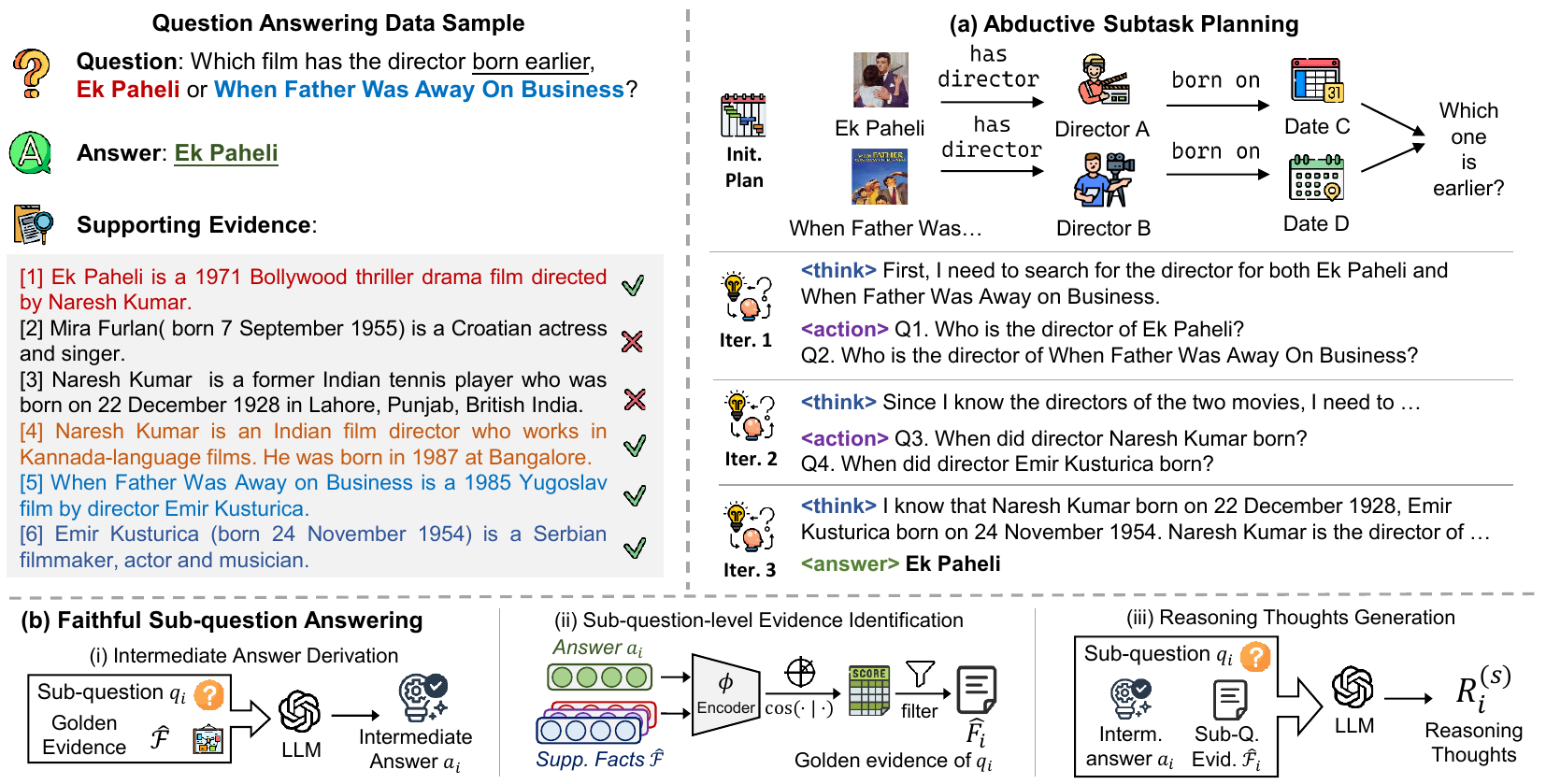}
	\caption{The EviPath framework. (a) Abductive Subtask Planning (top-right) initializes a plan and iteratively unfolds the planner's reasoning trajectory. (b) Faithful Sub-question Answering (bottom) derives grounded intermediate thoughts and answers using a proxy environment constructed from supporting evidence.}
	\label{fig:EviPath}
\end{figure*}

\subsection{Reasoning Path Synthesis as an Abductive Reasoning Task}
We consider an MHQA dataset to consist of training problems in the form of $d_{\text{train}} = (q, a, \mathcal{F}, \hat{\mathcal{F}})$, where $q$ is the question, $a$ is the answer, $\mathcal{F}$ represents the pool of supporting facts with distractors, and $\hat{\mathcal{F}}$ is the set of golden evidence satisfying $\hat{\mathcal{F}} \subset\mathcal{F}$.
We argue that \textbf{\textit{reasoning path synthesis}} constitutes an \textbf{\textit{abductive reasoning}}~\citep{josephson1996abductive} task. Given the golden evidence $\hat{\mathcal{F}}$ as the premise, we seek to infer a trajectory $\mathcal{T}_{q\rightarrow a}$ that best derives the observed answer $a$ from question $q$. Formally, we have the objective function:
\begin{align}
    \mathcal{T}_{q\rightarrow a}&=(\mathcal{P}, \{\mathcal{R}_i\}_{i=1}^{|\mathcal{P}|}, \{Q_i\}_{i=1}^{|\mathcal{P}|}, \{A_i\}_{i=1}^{|\mathcal{P}|}, \{\mathcal{R}_j^{(s)}\}_{j=1}^{N}) \notag \\ 
    &= f_{\text{LLM}}(q \wedge a | \hat{\mathcal{F}}),
\end{align}
where $A_i=\{a_j|q_j\in Q_i\}$ is the set of answers to the sub-questions in step $i$, $N=\sum_{i}|Q_i|$ is the total number of sub-questions to be resolved, $f_{\text{LLM}}$ denotes the LLM. Since $\hat{\mathcal{F}}$ provides the minimally sufficient ground truth evidence to answer question $q$, employing an abductive reasoning process based on $\hat{\mathcal{F}}$ yields concise trajectories. This approach naturally minimizes the number of reasoning steps while maintaining strict alignment with the expected outcome $a$.

\section{Method}\label{method}
In this section, we present EviPath, a reasoning-path synthesis framework aligned with the planner–executor architecture of RAG agents. The pipeline comprises two phases (Figure~\ref{fig:EviPath}): \textit{(i) Abductive Subtask Planning (ASP)} and \textit{(ii) Faithful Sub-question Answering (FSA)}, each of which corresponds to the Planner and Executor modules, respectively. We then leverage these complete reasoning paths to develop RAG agents via \textit{(iii) conversational fine-tuning~(CFT)}.

\subsection{Abductive Subtask Planning (Planner-side Reasoning Path Synthesis)}
\subsubsection{Task Decomposition} To solve a complex question $q$, the RAG agent first decomposes it into a plan $\mathcal{P}$ with a set of sub-questions $\{q_1, q_2, \cdots, q_n\}$. The quality of this initial plan is crucial, as it constrains the search space and improves overall accuracy and efficiency of reasoning. However, despite having strong semantic understanding capabilities, LLMs often fail to generate coherent multi-step plans without direct supervision. This challenge is compounded by the fact that mainstream QA datasets offer the final answer $a$, supporting facts $\mathcal{F}$, and a golden evidence subset $\hat{\mathcal{F}}$, but lack agent-oriented question decomposition for LLM fine-tuning. 
To bridge the aforementioned supervision gap, EviPath introduces \textbf{\textit{abductive reasoning}} to reverse-engineer a latent reasoning graph by analyzing the ground-truth answer and the dependencies between different pieces of evidence with an LLM.~\footnote{Note: For the MuSiQue dataset where logical sub-questions are available, we incorporate them as structural priors to synthesize the trajectory, where the executable plan with search actions and intermediate thoughts are derived via abductive reasoning.} The reasoning graph is then linearized into a concrete sequence of sub-questions, creating a ``golden'' plan that serves as an explicit supervision signal. Formally, the task decomposition process can be expressed as:
\begin{equation}
    \mathcal{P} = \{q_1, q_2, \cdots, q_n\} = \left\{f_{\text{TD}}(q_{<i}, q, a, \hat{\mathcal{F}})\right\}_{i=1}^{n}.
\end{equation}
It should be noted that sub-questions in the initial plan may be under-specified. These incomplete questions will be dynamically grounded and refined during plan execution, as answers from preceding steps provide the necessary context (e.g., entities, constraints) for subsequent sub-questions.

\subsubsection{Iterative Exploration} Upon obtaining the initial plan,  EviPath generates the solution by iteratively simulating an agent's task problem-solving process. Each iteration consists of two primary steps: \textit{think} and \textit{action}.

\paragraph{Think.} In this step, the planner of an RAG agent reviews answers to sub-questions resolved in preceding iterations $A_{i-1}$ and identifies the set of remaining sub-questions that are both solvable and essential to pursue in the current iteration. It then generates its internal monologue, or ``thoughts'' $\mathcal{R}_i$, enclosed within \texttt{<think>} and \texttt{</think>} tags. These thoughts detail: (i)~the instantiation of previously underspecified variables, (ii) a prioritized set of sub-goals for the current step,  and (iii) the resulting updates to the previous plan. %, and the dependencies among sub-questions. 
To align with the target of the original question, the thought generation process is conditioned on the current agent state, $s_i=\left\{\mathcal{P}, \{\mathcal{R}_j\}_{j=1}^{i-1}, \{Q_j\}_{j=1}^{i-1}, \{A_j\}_{j=1}^{i-1}\right\}$
, along with the final answer $a$, and golden evidence set $\mathcal{\hat{\mathcal{F}}}$. This step is formulated as:
\begin{equation}
    \mathcal{R}_i = f_{\text{think}}(s_i, a, \hat{\mathcal{F}}).
\end{equation}

\paragraph{Action.} In this step, the planner translates the priorized sub-goal(s) from its thought into concrete, executable retrieval queries $Q_i$. Specifically, the retrieval intent is explicitly rendered within \texttt{<action>} and \texttt{</action>} tags. Each retrieval query is written as a complete sub-question $q_j \in Q_i$ that can be executed independently. Similarly, let $m_i$ be the number of sub-questions that need to be solved in the $i$-th step, we have the objective function:
\begin{equation}
    Q_i = \{q_1, q_2, \cdots, q_{m_i}\} = \big\{f_{\text{action}}(\mathcal{R}_i,s_i, a, \hat{\mathcal{F}})\big\}_{j=1}^{m_i}.
\end{equation}

\subsection{Faithful Sub-question Answering (Executor-side Reasoning Path Synthesis)} 
After the planning step generates a sub-question $q_i$, we synthesize the corresponding reasoning path for the executor. This involves generating a chain-of-thoughts~$\mathcal{R}_i^{(s)}$ that processes the sub-question and its retrieved context to yield an intermediate answer $a_i$.

\paragraph{The challenge of real-time retrieval. } In practice, the executor of a RAG agent retrieves the relevant context of the sub-question from an external knowledge base. However, existing dense or sparse retrievers often fail to secure the necessary golden evidence. The imperfect retrieval is particularly problematic since mainstream MHQA datasets do not provide intermediate answers at the sub-question level. Without such granular supervision, any disruption to the evidence chain prevents the LLM from assembling a coherent reasoning path, and ultimately leads to unfaithful answers.

\paragraph{Robust trajectory synthesis in a simulated environment. }
To circumvent the aforementioned challenge, EviPath forgoes real-time retrieval and instead constructs a simulated environment for robust data synthesis. By utilizing the complete set of supporting facts $\mathcal{F}$ as a stable, local knowledge base, we ensure all necessary golden evidence for each sub-question is readily accessible, creating an ideal setting for generating high-fidelity reasoning paths.

Within this simulated environment, we synthesize the reasoning path for each sub-question $q_i$ through a three-step procedure:
\begin{enumerate}[label=(\roman*),leftmargin=24pt, nosep]
\item First, we provide the LLM with the sub-question $q_i$ and all golden evidence $\hat{\mathcal{F}}$ to \textbf{\textit{derive the intermediate answer}}, $a_i$. 
\item Then, we \textbf{\textit{identify the golden evidence set}} $\hat{\mathcal{F}}_i$\textbf{\textit{ for the current sub-question}} $q_i$. Specifically, we employ a sentence transformer to encode the complete answer sentence and every piece of golden evidence in $\hat{\mathcal{F}}$. All evidence having a cosine similarity to the answer that is higher than a threshold $\tau$ will be included in $\hat{\mathcal{F}_i}$.~\footnote{If none of the evidence satisfies the condition, we pick the one with the highest cosine similarity.} 
\item Finally, following the abductive reasoning paradigm, we task the LLM to \textbf{\textit{generate a chain-of-thought}} $\mathcal{R}_i^{(s)}$ that begins with sub-question $q_i$, identifies the set of key evidence $\hat{\mathcal{F}}_i$ from the noisy supporting facts $\mathcal{F}$, and culminates in the answer $a_i$. Considering that real-world retrieval results also contain substantial noise, leveraging $\mathcal{F}$ rather than $\hat{\mathcal{F}}$ as the pseudo retrieval context can better equip LLMs with better in-context reasoning ability. 
\end{enumerate}

Formally, the aforementioned process can be formulated as:
\begin{align}
a_i &= f_{\text{QA}}(q_i, \hat{\mathcal{F}}), \notag \\
\hat{\mathcal{F}}_i &= \{\zeta \in \hat{\mathcal{F}} \mid \cos(\phi(\zeta), \phi(a_i)) > \tau\}, \\
\mathcal{R}_i^{(s)} &= f_{\text{think}}^{(s)}(q_i, a_i, \hat{\mathcal{F}}_i,\mathcal{F}), \notag
\end{align}
where $\phi(\cdot)$ denotes the embedding encoded by text encoder $\phi$.

The reasoning path synthesis process continues until the final answer $a$ is reached. At this final step, the LLM refrains from issuing further \texttt{<action>} tags and concludes the trajectory by extracting the answer and wrapping it in \texttt{<answer>} and \texttt{</answer>} tags.

\begin{figure}[t]
    \centering
    \caption{Distribution of Reasoning Primitives across Different Question Types.}
    \includegraphics[width=1.0\linewidth]{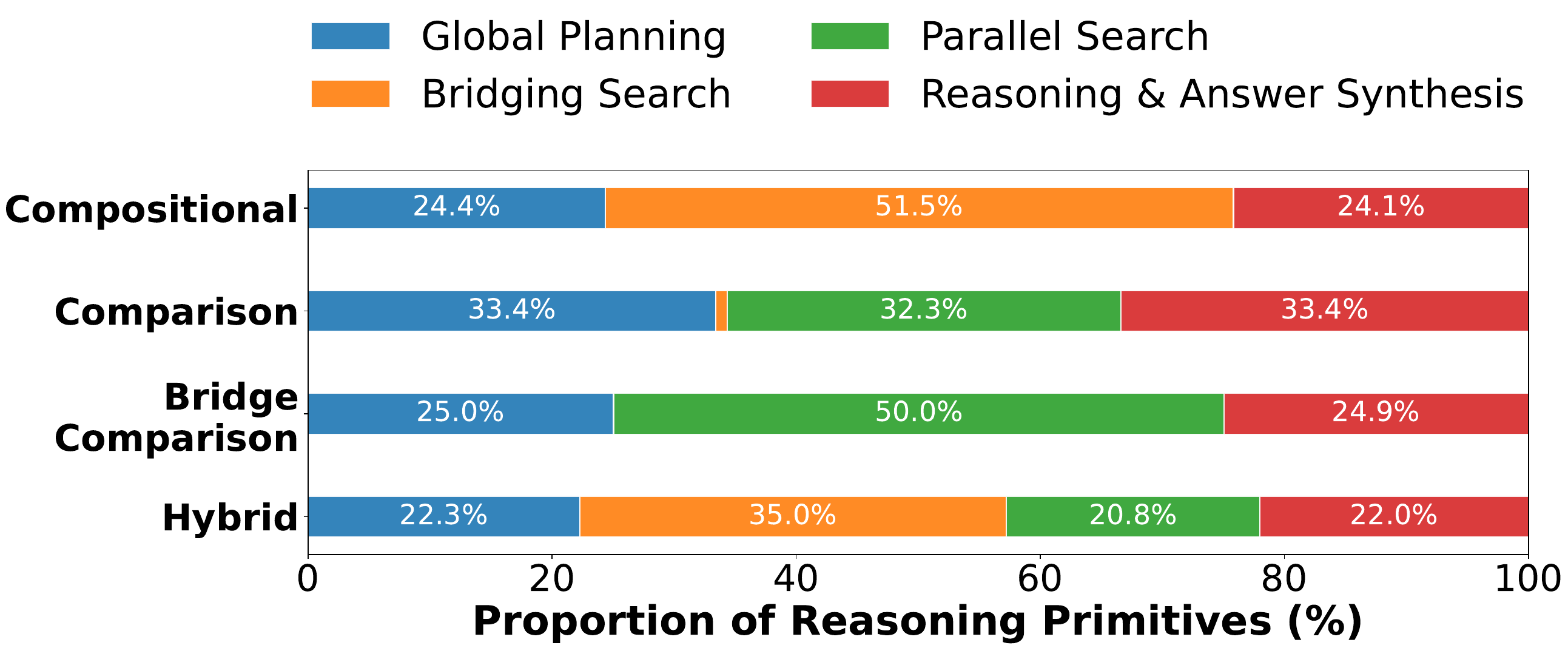}
    \label{fig:question_type_dist}
\end{figure}

\begin{table*}[t]
\centering
\small
\caption{Experiment results on multi-hop question answering benchmark datasets. 
The performance of vanilla CoT, RAG, and DecomP (with $^*$) are referred from~\cite{Collab-RAG}. Results for other baselines are taken from \underline{original research papers}. 
}
\begin{tabular}{lccccccccc}
\toprule
\multicolumn{1}{l}{\multirow{2.5}{*}{\textbf{Method}}}  &        \multicolumn{1}{c}{\multirow{2.5}{*}{\textbf{Backbone}}}          & \multicolumn{2}{c}{\textbf{HotpotQA}}  & \multicolumn{2}{c}{\textbf{MuSiQue}}   & \multicolumn{2}{c}{\textbf{2WikiMQA}}  & \multicolumn{2}{c}{\textbf{Average}}   \\
\cmidrule(lr){3-4} \cmidrule(lr){5-6} \cmidrule(lr){7-8}  \cmidrule(lr){9-10}
                  &                     & EM            & F1            & EM            & F1            & EM            & F1            & EM            & F1            \\
\midrule
CoT$^*$                     & GPT-4o                 & 29.4          & 48.9          & 17.0          & 28.9          & 41.8          & 53.6          & 29.4          & 43.8          \\
RAG$^*$                     & GPT-4o                 & 47.2          & 63.6          & 17.4          & 30.1          & 45.8          & 57.1          & 36.8          & 50.3         \\
DecomP$^*$                  & GPT-4o                 & 52.2          & 65.6          & 27.8          & 42.3          & 62.2          & 73.3          & 47.4          & 60.4          \\
RAFT                    & Llama3.1-8B            & 41.0          & 51.6          & 13.8          & 24.0          & 39.4          & 45.8          & 31.4          & 40.5          \\
RaFe                    & GPT4o-mini             & 40.6          & 55.4          & 12.4          & 25.3          & 36.2          & 39.3          & 29.7          & 40.0            \\
Iter-RetGen             & Instruct-GPT           & 45.1          & 60.4          & 26.1          & 42.0          & 50.2          & 65.3          & 40.5          & 55.9          \\
HippoRAG                & GPT3.5                 & 41.8          & 55.0          & 19.2          & 29.8          & 46.6          & 59.5          & 35.9          & 48.1          \\
IRCoT                   & GPT3                   & 49.3          & 60.7          & 26.5          & 36.5          & 57.7          & 68.0          & 44.5          & 55.1          \\
RQ-RAG                  & GPT4o-mini             & 46.4          & 59.4          & -             & -             & 50.2          & 58.8          & -             & -             \\
ReSP                    & Llama3-8B              & 47.2          & -             & -             & -             & 38.3          & -             & -             & -             \\
IterDRAG                & Gemini 1.5             & 38.4          & 49.8          & 22.6          & 35.0          & 44.3          & 54.6          & 35.1          & 46.5          \\
EfficientRAG            & Llama3-8B              & 50.6          & 57.9          & 16.4          & 21.2          & 44.2          & 51.6          & 37.1         & 43.6          \\
\midrule
\textit{RAG Agents} & & & & & & & & & \\
Search-o1               & QwQ 32B                & 45.2          & 57.3          & 16.6          & 28.2          & 58.0          & 71.4          & 39.9          & 52.3          \\
RAG-Gym                 & Llama3.1-8B            & 44.1          & 56.8          & -             & -             & 50.2          & 57.9          & -             & -             \\
Search-R1               & Qwen2.5-7B             & 43.3          & -             & 19.6          & -             & 38.2          & -             & 33.7          & -             \\
R1-Searcher             & Qwen2.5-7B             & -           & 60.4          & -             & 35.7             & -            & 62.8          & -             & 53.0             \\
Collab-RAG              & Llama3.1-8B & \underline{53.0}          & 65.6          & 26.4          & 42.4          & 63.2          & 74.6          & 47.5          & 60.9          \\
RAG-Star                & GPT4o-mini             & 46.0          & 60.0          & 22.2          & 30.7          & 38.0          & 46.8          & 35.4          & 45.8          \\
Mujica-MyGO             & Qwen2.5-7B             & 41.5          & 53.8          & 26.1          & 35.9          & 77.6          & 84.2          & 48.4          & 58.0          \\
R1-Searcher++           & Qwen2.5-7B             & -             & 59.0          & -             & 33.8          & -             & 61.2          & -             & 51.3          \\
DynaSearcher            & Qwen2.5-7B             & 52.0          & \underline{66.1}          & 26.5          & 38.7          & 61.9          & 72.0          & 46.8          & 58.9          \\
KG-o1                   & Llama3.1-8B            & 43.4          & 60.2          & -             & -             & 55.0            & 68.6          & -             & -             \\
% KG-o1                   & Qwen2.5-14B            & 51.0          & 67.1          & -             & -             & 62.4          & 74.5          & -             & -             \\
ESA-KGR                 & Qwen2.5-7B             & 36.8          & 47.3          & 10.5          & 18.0            & 49.5          & 58.1          & 32.3          & 41.1          \\
Graph-R1                & Qwen2.5-7B             & -             & 62.7          & -             & 46.2          & -             & 65.0          & -             & 58.0          \\
\midrule
\multirow{4}{*}{\textbf{EviPath (full)}} & Qwen2.5-7B & 51.3 & 64.0 & \underline{40.2} & \underline{50.0} & \textbf{92.0} & \textbf{94.3} & \underline{61.2} & \underline{69.4} \\
& Llama3.2-1B & 39.4 & 50.6 & 29.7 & 37.9 & 76.7 & 79.0 & 48.6 & 55.8 \\
& Llama3.2-3B & 48.6 & 60.7 & 39.9 & 48.8 & 90.4 & 92.9 & 59.6 & 67.4 \\
 & Llama3.1-8B            & \textbf{53.8} & \textbf{66.4} & \textbf{44.3} & \textbf{54.6} & \underline{91.3} & \underline{93.6} & \textbf{63.1} & \textbf{71.5} \\
\bottomrule
\end{tabular}
\label{tab:results}
\end{table*}

\subsection{Conversational Fine-tuning}
We generated \textbf{265k} reasoning trajectories using the \texttt{LLaMA\allowbreak 3.1\allowbreak-70B} model with few-shot demonstrations on the training splits of three multi-hop QA datasets.\footnote{We design and leverage different seed examples for different types of questions to better cover diverse reasoning patterns.}
Specifically, EviPath produces approximately \textbf{159k}, \textbf{86k}, and \textbf{20k} reasoning trajectories on 2WikiMultihopQA~\citep{2Wiki}, HotpotQA~\citep{HotpotQA}, and MuSiQue~\citep{MuSiQue} datasets, respectively.\footnote{The reported number of trajectories refers to the successfully synthesized planner reasoning paths after validity filtering.}
Each trajectory was then formatted to align with the architecture of the RAG agent, yielding one multi-turn \textit{Planner Prompt} for training complex high-level planning capabilities and multiple single-turn \textit{Executor Prompts} for training evidence-grounded sub-question answering. The data from the planner and executor sides are aggregated together in a unified supervised fine-tuning (SFT) process. Formally, the LLM is optimized by maximizing the following joint objective function: 
\begin{align}
\mathcal{J}_{\text{SFT}}(\theta)
&= \mathbb{E}_{(q,a,\mathcal{F})\sim\mathcal{D}_{\text{train}}}\Big[
  \pi_\theta(\mathcal{P}\mid q, \mathcal{I}_p) \notag \\
  &\cdot \prod_{i=1}^{|\mathcal{P}|-1} \pi_\theta(Q_i,\mathcal{R}_i \mid \mathcal{P},Q_{<i},a_{<i},q, \mathcal{I}_p) \notag
\\[-1pt] 
&\cdot \Big(\prod_{j=1}^{|Q_i|} \pi_\theta(a_j,\mathcal{R}_j^{(s)}\mid q_j,\mathcal{F}, \mathcal{I}_e)\Big)
  \; \notag \\ 
  &\cdot\; \pi_\theta\big(a,\mathcal{R}_{|\mathcal{P}|} \mid \mathcal{P},Q_{<|\mathcal{P}|-1},a_{<|\mathcal{P}|-1},q,a,\mathcal{I}_p\big)
\Big],
\end{align}

where $\mathcal{D}_{\text{train}}$ denotes the training dataset, $\pi_\theta$ is the policy of the backbone LLM with trainable parameters $\theta$, $\mathcal{I}_p$ and $\mathcal{I}_e$ are instruction prompts for the planner and the executor, respectively. 

\section{Experiments}
\subsection{Datasets, Baselines and Evaluation Metrics}
We conduct our main experiments on three multi-hop QA datasets, including text-based benchmarks HotpotQA~\citep{HotpotQA} and MuSiQue~\citep{MuSiQue} and knowledge graph-based question answering~(KBQA) benchmark 2WikiMultihopQA~(2WikiMQA)~\citep{2Wiki}. %The list of baseline methods are provided in Appendix~\ref{app:baselines}.
We compare EviPath with a comprehensive set of $24$ baselines, including vanilla CoT~\citep{CoT} and RAG~\citep{RAG}, iterative RAG pipelines such as DecomP~\citep{DecomP}, RAFT~\citep{RAFT}, RaFe~\citep{RaFe}, Iter-RetGen~\citep{Iter-RetRAG}, HippoRAG~\citep{HippoRAG}, IRCoT~\citep{IRCoT}, RQ-RAG~\citep{RQ-RAG}, ReSP~\citep{ReSP}, IterDRAG~\citep{IterDRAG}, and EfficientRAG~\citep{EfficientRAG}, RAG agents such as Search-o1~\citep{Search-o1}, RAG-Gym~\citep{RAG-Gym}, Search-R1~\citep{Search-R1}, Collab-RAG~\citep{Collab-RAG}, RAG-Star~\citep{RAG-Star}, R1-Searcher~\citep{R1-Searcher}, and Mujica-MyGO~\citep{Mujica-MyGO}, and concurrent methods such as R1-Searcher++~\citep{R1-Searcher++}, DynaSearcher~\citep{DynaSearcher}, KG-o1~\citep{KG-o1}, ESA-KGR~\citep{ESA-KGR}, and Graph-R1~\citep{Graph-R1}.
We evaluate EviPath and all baselines using Exact Match~(EM) and F1.
Detailed statistics of datasets are provided in Appendix~\ref{app:dataset}.

\subsection{Implementation}
We examine the effectiveness of our proposed method by fine-tuning four instruction-tuned LLMs with different scales: \texttt{Qwen2.5-}\allowbreak\texttt{7B}, \texttt{LLaMA 3.2-1B}, \texttt{LLaMA 3.2-3B}, and \texttt{LLaMA 3.1-8B}. 

During evaluation, we adopt the ``\textbf{open-domain}'' setting, where \textbf{\textit{the agent is required to retrieve relevant information from the external environment}}, while disregarding the supporting facts provided with dev. set questions. For HotpotQA, we use the official Wikipedia dump associated with the dataset as the retrieval corpus. For MuSiQue, following the convention~\citep{IRCoT}, we construct a large retrieval corpus by aggregating all supporting passages associated with each question. For 2WikiMQA and QALD-10, we use the official Web APIs provided by Wikidata~\citep{Wikidata}. We adopt \texttt{bge-large-en-v1.5} as the retriever in all of our experiments. 
More details for LLM fine-tuning are in Appendix~\ref{app:implementation}. \footnote{Our code and prompts are available at \url{https://github.com/RaynorLEE/EviPath}.}

\subsection{Analysis of the Synthetic Corpus}
\begin{table}[t]
\centering
\caption{Statistics of the synthetic corpus across datasets and question types. The success rate denotes the fraction of training data with valid planner prompts.}
\label{tab:full_stats}
\small
\setlength{\tabcolsep}{2.5pt}
\begin{tabular}{@{}llrrr@{}}
\toprule
\textbf{Dataset} & \textbf{Type} & \textbf{\#Question} & \textbf{\#Planner prompts} & \textbf{Succ. Rate(\%)} \\
\midrule
\multirow{3}{*}{HotpotQA}
 & Bridge & 72,991 & 69,841 & 95.7 \\
 & Comp. & 17,456 & 16,636 & 95.3 \\
 \cmidrule(l){2-5}
 & \textbf{Total} & \textbf{90,447} & \textbf{86,477} & \textbf{95.6} \\
\midrule
\multirow{4}{*}{MuSiQue}
 & 2-hop & 14,376 & 12,916 & 89.8 \\
 & 3-hop & 4,387 & 3,958 & 90.2 \\
 & 4-hop & 1,175 & 1,109 & 94.4 \\
 \cmidrule(l){2-5}
 & \textbf{Total} & \textbf{19,938} & \textbf{17,983} & \textbf{90.2} \\
\midrule
\multirow{5}{*}{2WikiMQA}
 & B.C. & 34,631 & 33,506 & 96.6 \\
 & Comp. & 51,963 & 50,354 & 96.9 \\
 & Compos. & 76,481 & 72,565 & 94.9 \\
 & Infer. & 4,378 & 4,187 & 95.6 \\
 \cmidrule(l){2-5}
 & \textbf{Total} & \textbf{167,453} & \textbf{160,612} & \textbf{95.9} \\
\bottomrule
\end{tabular}
\end{table}
Figure~\ref{fig:question_type_dist} illustrates the structural adaptivity of the EviPath corpus, where the reasoning primitives distribution of synthetic trajectories shifts according to the task's intrinsic logic. Specifically, Parallel Search (Green) is activated for comparison and bridge-comparison questions, effectively handling independent subtasks that require parallel search actions. In contrast, Bridging Search (Orange) dominates compositional tasks to support sequential evidence chaining. This alignment confirms that our abductive synthesis framework allows LLMs to accurately reconstruct the underlying dependency graph, providing precise and high-quality supervision signals for agent development. The high success rate in Table~\ref{tab:full_stats} further shows that EviPath successfully mitigates the stochastic failures common in open-ended exploration, ensuring the stable production of high-fidelity training trajectories for open-domain question answering. 

\subsection{Main Results}
The experiment results in Table~\ref{tab:results} demonstrate that EviPath is a simple yet effective scheme for synthesizing reasoning trajectories for training RAG agents. Despite relying solely on SFT, the 8B model trained on our synthetic trajectories significantly outperforms all baselines, including those leveraging large-scale LLMs~(e.g. GPT-4o) or complex RL algorithms~(e.g. GRPO~\citep{shao2024deepseekmath}), achieving an average absolute EM gain of \textbf{14.7\%}. The substantial improvement reaffirms persistent data limitations in RAG agent training and highlights the importance of introducing precise, evidence-anchored reasoning paths. 
Our EviPath-trained agents exhibit a clear scaling effect, with larger backbone LLMs consistently improving QA performance. \textbf{\textit{More importantly, our results demonstrate that the process-supervised trajectories can offset model sizes, enabling smaller LLMs to overcome their limited reasoning capabilities.}} Specifically, RAG agents equipped with 1B and 3B \texttt{LLaMA 3.2} models trained on our synthetic data substantially surpass all baseline models on the 2WikiMQA and MuSiQue datasets.

EviPath excels on both text-based and knowledge graph-based QA. Its strong performance on 2WikiMQA demonstrates its ability to leverage structured knowledge graphs to capture logical dependencies between sub-questions. Unlike other graph-based baselines~(e.g., Graph-R1, KG-o1), EviPath's evidence-anchored supervision compels agents to remain faithful to the graph structure, thereby encouraging the selection of optimal reasoning paths.
 
The modest gains on HotpotQA stem from its simpler 2-hop questions, which place fewer demands on reasoning capabilities for LLMs.
In contrast, EviPath excels on the complex MuSiQue dataset with 2-to-4 hop questions, significantly outperforming RL baselines like Search-R1~\citep{Search-R1}. This highlights a key limitation of RL: without a foundational ability to solve a problem, an agent cannot acquire the positive rewards needed for self-improvement. To summarize,  \textbf{\textit{the primary barrier to powerful RAG agents is the scarcity of high-quality, process-level supervision signals, rather than model scale or learning algorithms.}}

\begin{table*}[t]
    \centering
    \small

    % ==================== 1. Ablation 表格 ====================
    \caption{Ablation study results of EviPath components on multi-hop question answering datasets.}
    \begin{tabular}{lccccccccc}
    \toprule
    \multicolumn{1}{l}{\multirow{2.5}{*}{\textbf{Method}}} & \multicolumn{1}{c}{\multirow{2.5}{*}{\textbf{Backbone}}} & \multicolumn{2}{c}{\textbf{HotpotQA}} & \multicolumn{2}{c}{\textbf{MuSiQue}} & \multicolumn{2}{c}{\textbf{2WikiMQA}} & \multicolumn{2}{c}{\textbf{Average}} \\
    \cmidrule(lr){3-4} \cmidrule(lr){5-6} \cmidrule(lr){7-8} \cmidrule(lr){9-10}
     & & EM & F1 & EM & F1 & EM & F1 & EM & F1 \\
    \midrule
    \textbf{EviPath (full)} & \textbf{Llama3.1-8B} & \textbf{53.8} & \textbf{66.4} & \textbf{44.3} & \textbf{54.6} & \textbf{91.3} & \textbf{93.6} & \textbf{63.1} & \textbf{71.5} \\
    - w/ pretrained LLM & Llama3.1-8B & 19.5 & 30.8 & 6.2 & 14.7 & 57.6 & 62.3 & 27.8 & 35.9 \\
    - w/ pretrained LLM & Llama3.1-70B & 31.0 & 44.9 & 13.1 & 23.4 & 84.6 & 87.7 & 42.9 & 52.0 \\
    - w/o planner fine-tuning & Llama3.1-8B & 41.6 & 54.3 & 27.6 & 37.7 & 45.2 & 50.4 & 38.1 & 47.5 \\
    - w/o executor fine-tuning & Llama3.1-8B & 48.9 & 61.8 & 31.2 & 41.9 & 86.1 & 91.6 & 55.4 & 65.1 \\
    - w/o supporting facts & Llama3.1-8B & 51.8 & 65.0 & 34.0 & 44.3 & 91.1 & 93.4 & 59.0 & 67.6 \\
    \bottomrule
    \end{tabular}
    \label{tab:ablation}

    \vspace{1.2em} 

    % ==================== 2. OOD 表格 ====================
    % 組列（Group column）
    \newcolumntype{G}{@{}>{\centering\arraybackslash}m{3.2cm}@{}}
    % 兩欄微表巨集
    \newcommand{\twoc}[2]{%
      \begingroup
      \setlength{\tabcolsep}{0.5pt}% 
      \renewcommand{\arraystretch}{1.0}%
      \begin{tabular}{@{}m{0.5\linewidth}<{\centering} m{0.5\linewidth}<{\centering}@{}}%
      #1 & #2%
      \end{tabular}%
      \endgroup
    }

    \caption{Experimental results in out-of-domain settings. Search-R1 does not support open-domain KBQA. Graph-R1 needs predefined textual corpus for graph construction, which is not provided in QALD-10.}
    \begin{tabular}{l l G G G}
    \toprule
    \multicolumn{2}{l}{\textbf{Datasets: (Training - Eval.)}} &
    \multicolumn{1}{c}{\textbf{HotpotQA - MuSiQue}} &
    \multicolumn{1}{c}{\textbf{MuSiQue - HotpotQA}} &
    \multicolumn{1}{c}{\textbf{2WikiMQA - QALD10}} \\
    \cmidrule(lr){3-3}\cmidrule(lr){4-4}\cmidrule(l){5-5}
    \textbf{Method} & \textbf{Backbone} &
    \twoc{EM}{F1} &
    \twoc{EM}{F1} &
    \twoc{EM}{F1} \\
    \midrule
    Search-R1   & Llama3.1-8B  & \twoc{12.3}{20.4} & \twoc{31.0}{42.3} & \twoc{N/A}{N/A} \\
    Graph-R1   & Qwen2.5-7B  & \twoc{25.0}{28.6} & \twoc{\textbf{42.9}}{50.0} & \twoc{N/A}{N/A} \\
    Mujica-MyGO & Qwen2.5-7B  & \twoc{26.1}{35.9} & \twoc{-}{-} & \twoc{39.9}{\textbf{49.7}} \\
    \textbf{EviPath} & Llama3.1-8B  & \twoc{\textbf{35.9}}{\textbf{46.3}} & \twoc{38.8}{\textbf{50.1}} & \twoc{\textbf{43.9}}{48.6} \\
    \bottomrule
    \end{tabular}
    \label{tab:ood}

    \vspace{1.2em}

    % ==================== 3. Generator 表格 ====================
    \caption{Question answering performance of RAG agents trained with reasoning paths synthesized by different LLMs.}
    \begin{tabular}{lcccccccccc}
    \toprule
    \multirow{2.5}{*}{\textbf{Method}} & \multirow{2.5}{*}{\textbf{\shortstack{Data Synthesis\\ LLM}}} & \multirow{2.5}{*}{\textbf{\shortstack{RAG Agent\\ Backbone}}} & \multicolumn{2}{c}{\textbf{HotpotQA}} & \multicolumn{2}{c}{\textbf{MuSiQue}} & \multicolumn{2}{c}{\textbf{2WikiMQA}} & \multicolumn{2}{c}{\textbf{Average}}        \\
    \cmidrule(lr){4-5} \cmidrule(lr){6-7} \cmidrule(lr){8-9}  \cmidrule(lr){10-11}
                                     &       &                                            & EM                & F1                & EM                & F1               & EM                & F1                & EM                   & F1                   \\
    \midrule
    \textbf{EviPath}                       & Llama3.1-8B & Llama3.1-8B                                       &       50.9            &        63.5           & 39.1              & 49.0             &         86.1          &        90.3           & \multicolumn{1}{l}{58.7} & \multicolumn{1}{l}{67.6} \\
    \textbf{EviPath}                 & Llama3.1-70B & Llama3.1-8B                                     & \textbf{53.8}     & \textbf{66.4}     & \textbf{44.3}     & \textbf{54.6}    & \textbf{91.3}     & \textbf{93.6}     & \textbf{63.1}        & \textbf{71.5}  \\
    \bottomrule
    \end{tabular}
    \label{tab:generator}

    \vspace{1.2em}

    % ==================== 4. Abductive 表格 ====================
    \caption{Performance comparison between deductive and abductive reasoning path synthesis. Here, ``pt.'' and ``ft.'' stand for pre-trained and fine-tuned models, respectively.}
    \begin{tabular}{lccccccc}
    \toprule
    \multirow{2.5}{*}{\textbf{Method}} & \multirow{2.5}{*}{\textbf{\shortstack{Data Synthesis\\LLM}}} & \multirow{2.5}{*}{\textbf{\shortstack{Data Synthesis \\Mode}}} & \multirow{2.5}{*}{\textbf{\shortstack{RAG Agent\\Backbone}}} & \multicolumn{2}{c}{\textbf{HotpotQA}} & \multicolumn{2}{c}{\textbf{2WikiMQA}} \\
    \cmidrule{5-6} \cmidrule{7-8} 
                                     &                                              &                                               &                                              & EM                & F1                & EM                & F1                \\
    \midrule
    \textbf{EviPath}                          & N/A                                          & N/A                                           & Llama3.1-8B (pt.)                             & 19.5              & 30.8              & 57.6              & 62.3              \\
    \textbf{EviPath}                         & Llama3.1-8B (pt.)                                  & Deductive                                     & Llama3.1-8B (ft.)                             & 39.7              & 51.2              & 80.3              & 85.7              \\
    \textbf{EviPath}                          & Llama3.1-8B (pt.)                                  & Abductive                                     & Llama3.1-8B (ft.)                             & \textbf{50.9}     & \textbf{63.5}     & \textbf{86.1}     & \textbf{90.3}   \\
    \bottomrule
    \end{tabular}
    \label{tab:abductive}

    \vspace{1.2em}

    % ==================== 5. Multiagent 表格 ====================
    \caption{Question answering performance comparison between different LLM deployment settings.}
    \begin{tabular}{lcccccccccc}
    \toprule
    \multirow{2.5}{*}{\textbf{Method}} & \multirow{2.5}{*}{\textbf{Deployment}} & \multirow{2.5}{*}{\textbf{Backbone}} & \multicolumn{2}{c}{\textbf{HotpotQA}} & \multicolumn{2}{c}{\textbf{MuSiQue}} & \multicolumn{2}{c}{\textbf{2WikiMQA}} & \multicolumn{2}{c}{\textbf{Average}} 
    \\
    \cmidrule(lr){4-5} \cmidrule(lr){6-7} \cmidrule(lr){8-9}  \cmidrule(lr){10-11}
                                     &  &                                    & EM                & F1                & EM                & F1               & EM                & F1                & EM                & F1               \\
    \midrule
    \textbf{EviPath}                 & Multiple LLMs      & 2$\times$Llama3.1-8B                  & 53.3              & 65.9              & 43.6              & 53.4             & 90.2              & 92.6              & 62.4              & 70.6             \\
    \textbf{EviPath}                 & Single LLM       & Llama3.1-8B                    & \textbf{53.8}     & \textbf{66.4}     & \textbf{44.3}     & \textbf{54.6}    & \textbf{91.3}     & \textbf{93.6}     & \textbf{62.8}     & \textbf{71.3}   \\
    \bottomrule
    \end{tabular}
    \label{tab:multiagent}

\end{table*}

\subsection{Ablation Studies}
We examine the effectiveness of EviPath in different settings by answering the following research questions~(RQs). 

\textbf{RQ1: Does the use of question-specific supporting evidence improve the quality of synthesized reasoning paths?}
We evaluated the necessity of supporting evidence by reconfiguring our data synthesis pipeline to use only question–answer pairs, compelling the LLM to retrieve relevant contexts from an external knowledge base and construct a complete reasoning path. As detailed in Table~\ref{tab:ablation}, the exclusion of supporting evidence resulted in performance degradation for our 8B LLM-based RAG agent across all three datasets. This performance drop is attributable to the loss of the implicit reasoning path implied in supporting evidence, which typically constrains the model's search space and ensures faithful derivations. In its absence, the model is vulnerable to two failure modes: (1) imperfect retrieval, where the inability to find ``golden'' evidence leads to plausible but incorrect reasoning, and (2) inherent limitations in the LLM's ability to reason about complex questions without explicit guidance. The latter issue is particularly acute on MuSiQue, which demands the composition of multiple facts and thus exhibits the most severe degradation.

\textbf{RQ2: Which core capability of LLMs is the primary limitation for building RAG agents?} To identify the primary limitation of LLMs in RAG agents, we trained two specialized models: a planner for high-level planning and an executor for sub-question answering. As shown in Table~\ref{tab:ablation}, replacing either specialized model with a pre-trained LLM degrades performance. Notably, this degradation is far more pronounced when replacing the planner. This result indicates that the primary bottleneck of LLM is not semantic understanding but long-horizon planning and reasoning, which reaffirms the critical need for high-quality reasoning trajectories in RAG agent development.

\textbf{RQ3: Can RAG agents trained on our offline synthetic data generalize to out-of-domain scenarios?} 
We evaluate out-of-domain~(OOD) generalization using cross-dataset transfer from HotpotQA to MuSiQue and vice versa (for textual QA), as well as from 2Wiki-KG to QALD-10 (for KBQA).~\footnote{QALD-10 does not have a training set. } As shown in Table~\ref{tab:ood}, models fine-tuned on our synthetic trajectories achieve comparable or superior transferability to state-of-the-art methods optimized with GRPO~\citep{shao2024deepseekmath}. Most notably, a model fine-tuned exclusively on 2-hop questions from HotpotQA shows remarkable generalization capabilities: it not only surpasses OOD baselines on MuSiQue (which features more complex 3–4 hop questions) but also outperforms all in-domain baselines trained directly on the MuSiQue dataset. 

\begin{figure*}[t]
    \centering
    \caption{Step-wise EM/F1 score on three open-domain QA datasets based on Qwen 2.5-7B model.}
    % hotpotqa
    \begin{subfigure}[t]{0.24\linewidth}
        \centering
        \includegraphics[width=\linewidth]{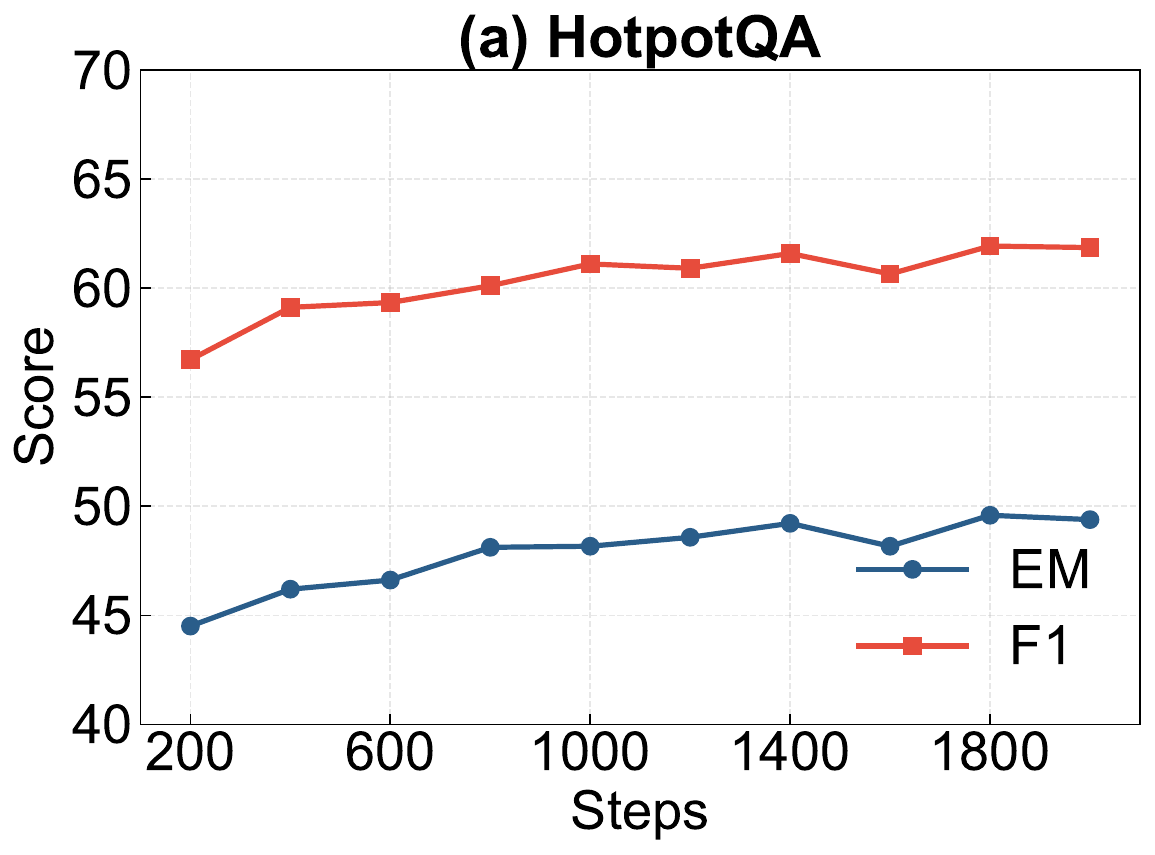}
        \label{fig:subfig_a}
    \end{subfigure}
    \hfill
    % musique
    \begin{subfigure}[t]{0.24\linewidth}
        \centering
        \includegraphics[width=\linewidth]{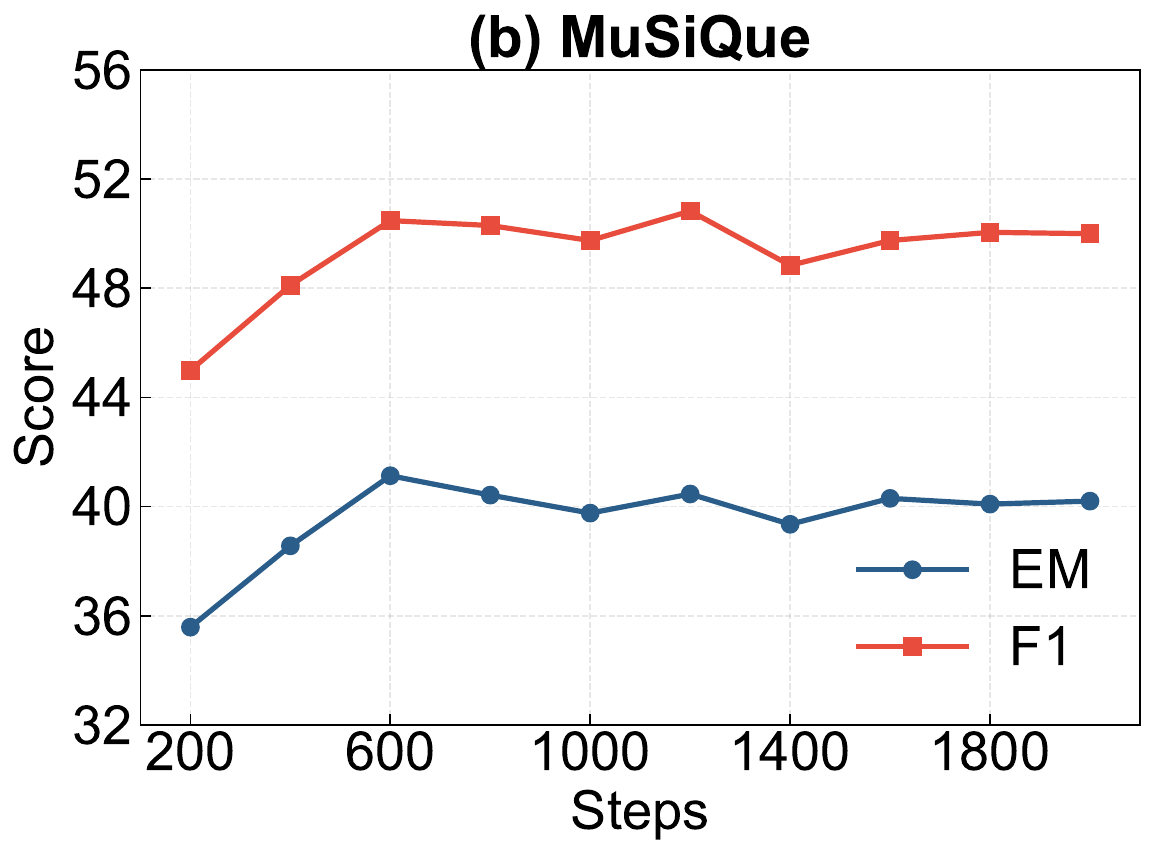}
        \label{fig:subfig_b}
    \end{subfigure}
    \hfill
    % 2wikimqa
    \begin{subfigure}[t]{0.24\linewidth}
        \centering
        \includegraphics[width=\linewidth]{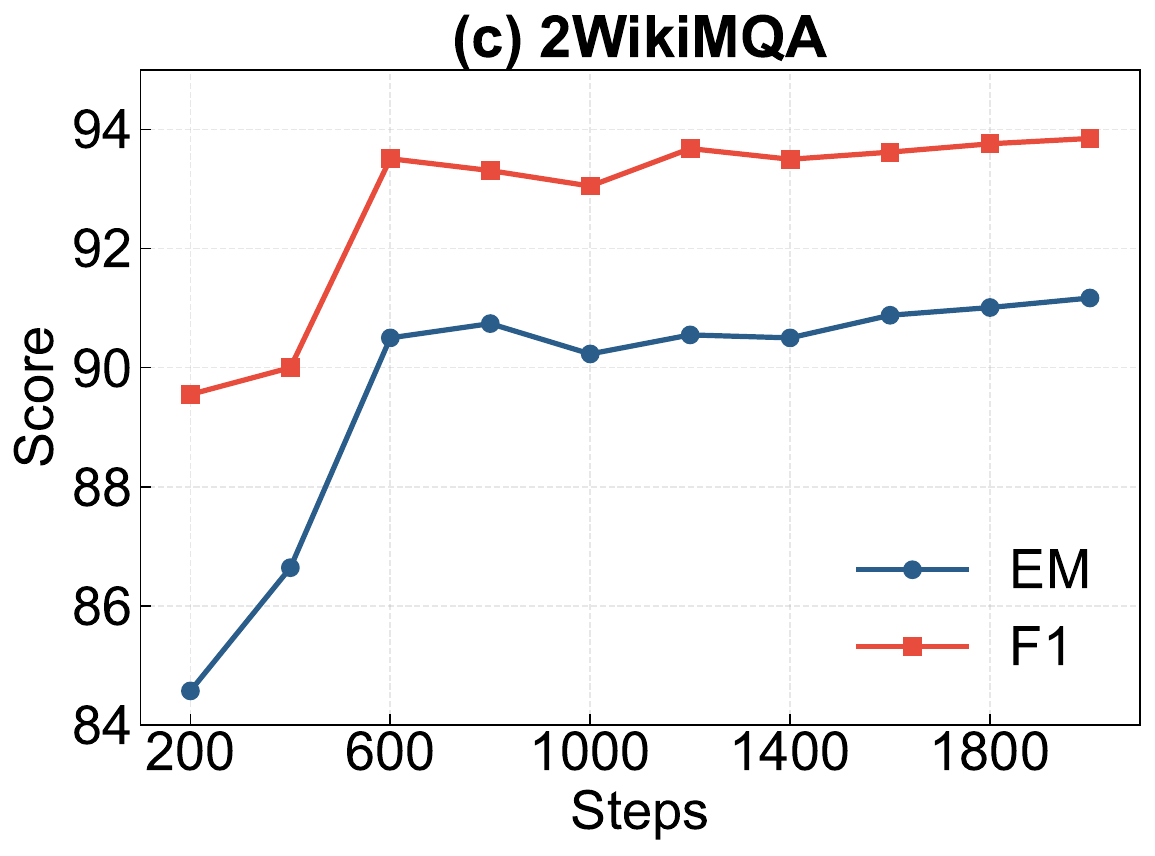}
        \label{fig:subfig_c}
    \end{subfigure}
    \hfill
    % average
    \begin{subfigure}[t]{0.24\linewidth}
        \centering
        \includegraphics[width=\linewidth]{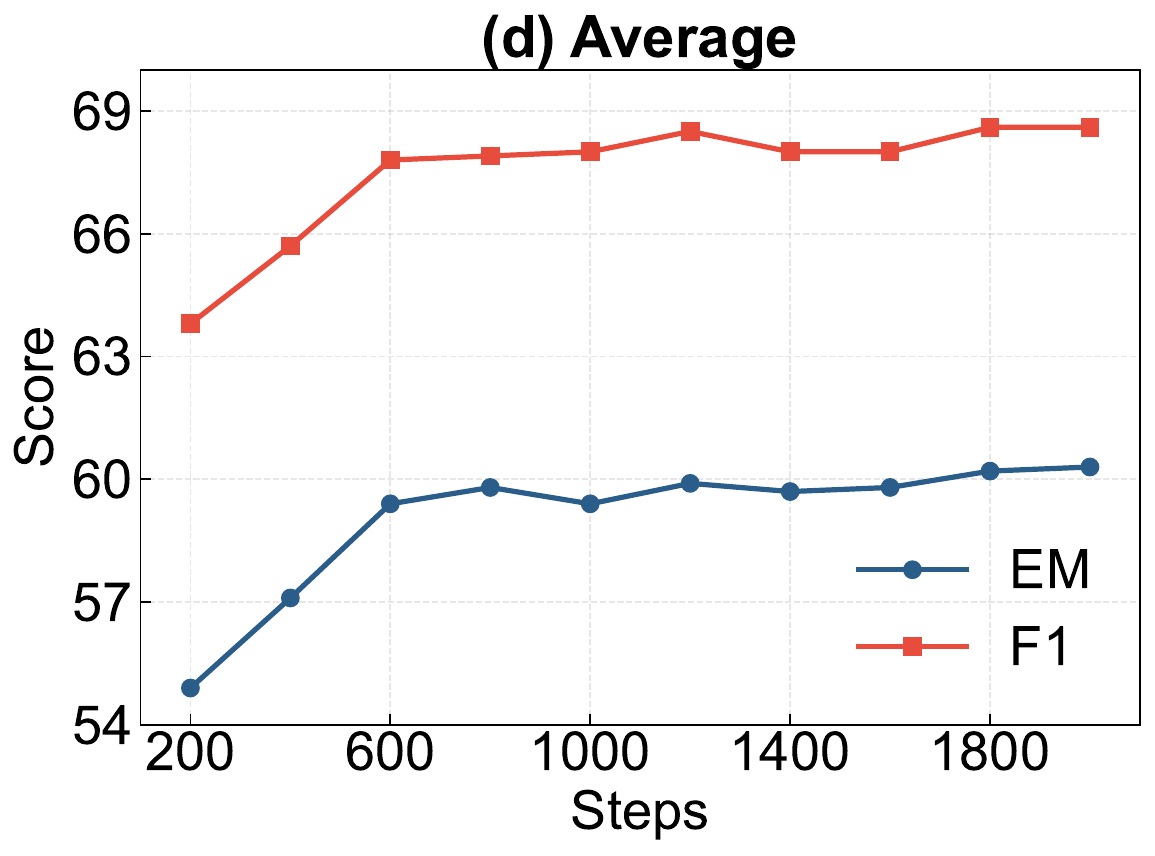}
        \label{fig:subfig_d}
    \end{subfigure}
    \label{fig:em_f1}
\end{figure*}

\begin{figure}[t]
    \centering
    \caption{End-to-end QA performance (in F1) with LLaMA-3.1-8B model under the distractor setting (N/A for KBQA).}
    \includegraphics[width=1.0\linewidth]{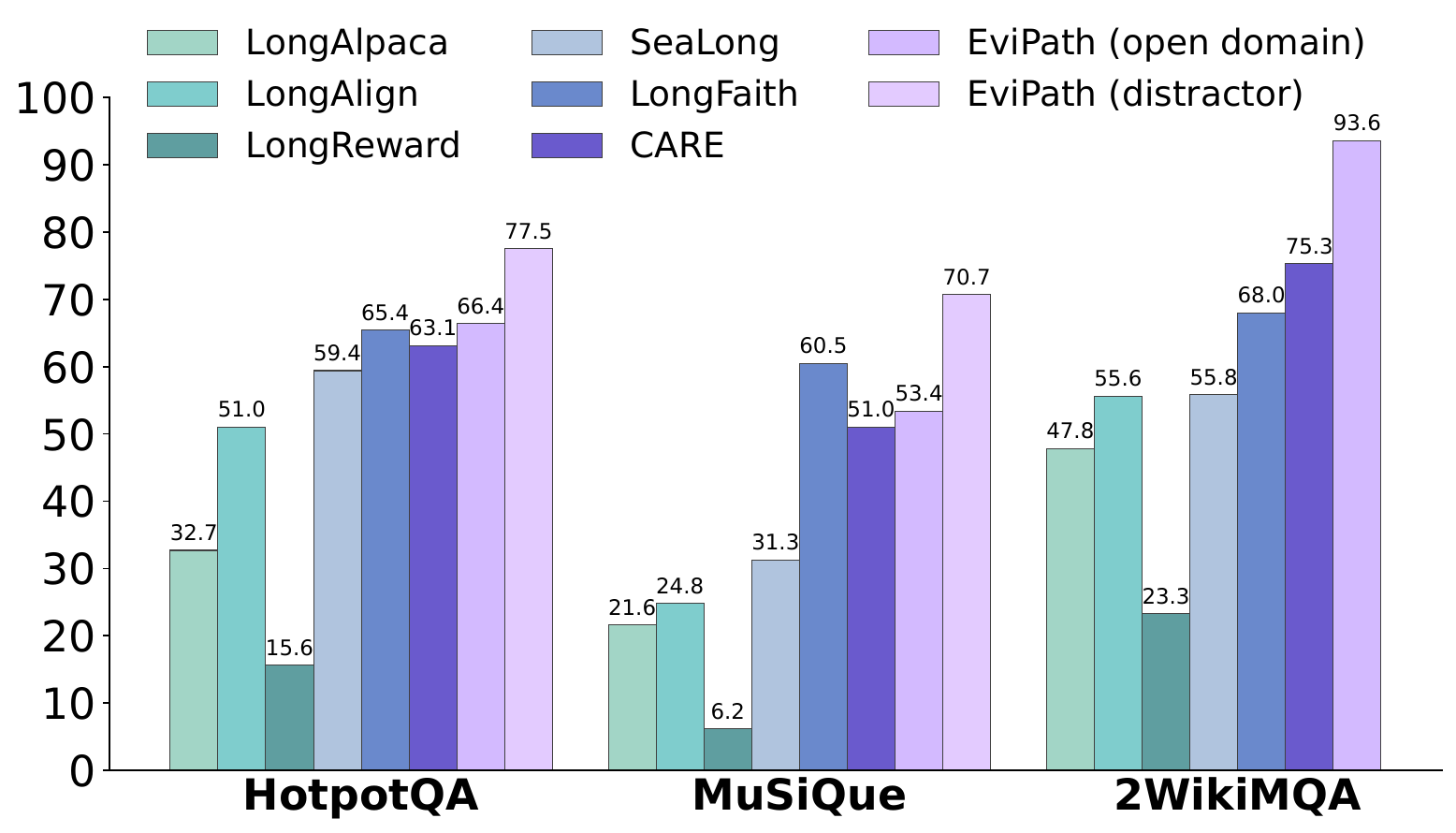}
    \label{fig:ood}
\end{figure}

\textbf{RQ4: To what extent does imperfect retrieval affect the end-to-end QA performance?} 
To isolate the impact of retrieval accuracy on end-to-end QA performance, we also evaluate EviPath-trained RAG agents in the distractor setting, where we use the 20 supporting facts~(paragraphs) provided with each test sample to simulate the retrieval results. From Figure~\ref{fig:ood} we observe that when the golden evidence is guaranteed to be included in the retrieval results, the performance ceiling rises substantially, underscoring the need for developing more advanced retrieval methods. In addition, the distractor setting allows us to make a direct comparison between EviPath and state-of-the-art reasoning path synthesis baselines, namely LongAlphaca~\citep{LongAlphaca}, LongAlign~\citep{LongAlign}, LongReward~\citep{LongReward}, SeaLong~\citep{SeaLong}, LongFaith~\citep{yang2025longfaith}, and CARE~\citep{CARE}. The double-digit gains demonstrate that while agentic RAG systems are not explicitly designed for the distractor setting, they still outperform non-agentic LLMs in long-context reasoning.

We further use the gap between the open-domain and distractor settings to analyze the root causes of end-to-end errors. Comparing F1 scores, imperfect retrieval leads to an $11.1$- and $17.3$- point performance drop on HotpotQA and MuSiQue, respectively. In contrast, even when golden evidence is guaranteed in the distractor setting, the remaining gaps to 100\% F1 are $22.5$ and $29.3$ points, respectively. These residual gaps correspond to non-retrieval errors, including failures in planning, reasoning, evidence aggregation, and answer generation. The larger residual gaps suggest that, although retrieval quality is important, long-horizon planning and reasoning remain the more pronounced bottleneck, especially on the more complex MuSiQue dataset with 2--4 reasoning hops.

\textbf{RQ5: How does the capacity of reasoning path generator affect the quality of synthetic reasoning paths and the final question answering performance?} To investigate the impact of the underlying LLM's capacity on the quality of EviPath synthesized trajectories, we generated a new set of training data using the Llama-3.1-8B model. From Table~\ref{tab:generator} we conclude that RAG agents trained on reasoning paths synthesized by the 70B Llama model yield better QA performance. However, an 8B model is already sufficient to synthesize high-quality reasoning paths that allow a RAG agent to attain state-of-the-art performance.%, demonstrating the robustness of the EviPath paradigm.

\textbf{RQ6: How does increasing the trajectories for LLM fine-tuning gradually improve QA performance? } 
Figure~\ref{fig:em_f1} shows that model performance scales with the volume of training data during the early stages of the SFT process, but the gains exhibit diminishing returns. 
Moreover, on large datasets like HotpotQA and 2Wiki, training for 2,000 steps (approx. \textbf{36,000 trajectories}) achieves performance within approximately 1\% of that from training on the full dataset. In contrast, the typical state-of-the-art RL-based method Graph-R1~\citep{Graph-R1} necessitates 30,720 examples from 6 datasets for 3 training epochs and samples 5 trajectories per question, resulting in roughly \textbf{460k trajectories} used to achieve the reported performance. This represents a significant efficiency advantage of EviPath over RL-based exploration, as it avoids the need for repetitive sampling.

\textbf{RQ7: To what extent does abductive reasoning improve the quality of synthetic reasoning paths? } To isolate the benefit of abductive reasoning, we re-synthesize the reasoning paths using a deductive, self-generation approach. In this setup, an Llama-3.1-8B model-based RAG agent is instructed to answer training questions without access to ground-truth answers or supporting facts. The results in Table~\ref{tab:abductive} show that SFT offers limited improvement when it primarily reinforces skills and knowledge that the LLM already possesses. In contrast, our abductive approach reverse-engineers paths from answers and supporting facts, which effectively lowers the dependency on model priors, unlocking a significantly higher performance ceiling.

\textbf{RQ8: Does deploying the Planner and the Executor modules of a RAG agent on two specialized LLMs outperform the same RAG agent supported by one single LLM?} The planner and executor modules address distinct aspects of the question answering task. To compare deployment strategies, we trained both a single LLM on all data and two specialized LLM on partitioned data for the planner and executor sides. As shown in Table~\ref{tab:multiagent}, the performance difference between the single-LLM and dual-LLM setups is modest. The slight advantage of the single-LLM setup suggests a positive transfer learning effect, where training on both planning and in-context reasoning tasks is mutually beneficial. This indicates that the key to unlocking the potential of small-scale LLMs is not specialization, but the quality of the training data.

\section{Conclusion}
In this paper, we introduce EviPath, a novel framework that uniquely applies abductive reasoning to reverse-engineer complete, evidence-anchored reasoning paths that include explicit task decomposition, retriever use, reasoning thoughts, and intermediate answers. EviPath overcomes the limitations of outcome-rewarded RL and static CoT trajectory synthesis methods, establishing an efficient, data-centric paradigm for RAG agent development. Experiments on commonly adopted open-domain QA benchmarks demonstrate that our synthetic data significantly boosts the in-domain accuracy and out-of-domain generalization capability of RAG agents. In the future, we plan to investigate the integration of process-supervised signals with policy gradient optimization methods and explore the potential of extending our data synthesis paradigm to other agentic tasks, such as mathematical reasoning and code generation.

\newpage

\bibliography{reference}
\bibliographystyle{ACM-Reference-Format}

\clearpage
% \onecolumn
\appendix
\section{Dataset Statistics}\label{app:dataset}
\begin{table}[h]
\centering
\small
\caption{Statistics of Datasets}
\begin{tabular}{lccccc}
\toprule
\textbf{Datasets}         & \textbf{\#Train} & \textbf{\#Dev} & \textbf{\#Test} & \textbf{\#Hops} & \textbf{Corpus}\\
\midrule
HotpotQA & 90447   & 7405  & 7405   & 2   & Text \\
MuSiQue  & 19938   & 2417  & 2459   & 2-4 & Text \\
2WikiMultihopQA & 167454  & 12576 & 12576  & 2-5 & Text and KG \\
QALD-10  & -       & 394   & -      & 1-2 & KG \\
\bottomrule
\end{tabular}
\label{tab:dataset}
\end{table}
We conduct our main experiments on three multi-hop QA datasets, including text-based benchmarks HotpotQA~\citep{HotpotQA} and MuSiQue~\citep{MuSiQue}, and KBQA benchmark 2WikiMultihopQA~\citep{2Wiki}. In our ablation studies, we also include another KBQA dataset QALD-10~\citep{QALD-10} for out-of-domain evaluation. Table~\ref{tab:dataset} shows the detailed statistics for all four datasets. 

\section{Implementation Details}
\label{app:implementation}
\begin{table}[h]\centering
\small
\caption{Hyperparameter settings.}
% \vspace{0.2cm}
% \resizebox{0.87\linewidth}{!}{
\begin{tabular}{lc}
\toprule
\textbf{Hyperparameters} & \textbf{Settings} \\ 
\midrule
Threshold $\tau$ & $0.9$ \\ 
SFT learning rate & 2e-6 \\ 
Per-device batch size & 2 \\ 
Gradient accumulation step & 8 \\ 
\# Epochs & $2$ \\
Warmup ratio & $0.1$ \\
Top-p & $0.8$ \\
LLM inference temperature & $0.7$ \\
Max. output tokens & $4096$ \\
Repetition penalty & $1.05$ \\
\bottomrule
\end{tabular}
% }
\label{tab:hyperparameters}
\end{table}
All of our experiments are conducted on 4 GPUs, each equipped with 80GB VRAM. We leverage the vLLM framework for accelerated inference during reasoning path synthesis, using a tensor parallel size of $4$.  We train all backbone LLMs with full-parameter fine-tuning using LLaMA-Factory~\citep{zheng2024llamafactory}. Specific hyperparameters are detailed in Table~\ref{tab:hyperparameters}. 

\begin{table*}[!b]
\centering
\caption{Comparison between EviPath and Pipeline RAG-based question answering solution. Results with * are referred from original research papers. Other results of baseline methods are reproduced with their publicly available implementations. }
\begin{tabular}{lcccccccc}
\toprule
   \multirow{2.5}{*}{\textbf{Method (Backbone / Setting)}}                           & \multicolumn{2}{c}{\textbf{HotpotQA}}  & \multicolumn{2}{c}{\textbf{MuSiQue}}   & \multicolumn{2}{c}{\textbf{2WikiMQA}}  & \multicolumn{2}{c}{\textbf{Average}}   \\
\cmidrule(lr){2-3} \cmidrule(lr){4-5} \cmidrule(lr){6-7} \cmidrule(lr){8-9}
                              & \textbf{EM}            & \textbf{F1}            & \textbf{EM}            & \textbf{F1}            & \textbf{EM}            & \textbf{F1}            & \textbf{EM}            & \textbf{F1}            \\
\midrule
IRCoT (GPT3)*                 & 49.3          & 60.7          & 26.5          & 36.5          & 57.7          & 68.0          & 44.5          & 55.1          \\
IRCoT (GPT-4o)                & 49.0          & 64.3          & 31.2          & 45.4    & 58.0          & 71.9          & 46.1          & 60.5          \\
HippoRAG (GPT3.5)*             & 41.8          & 55.0          & 19.2          & 29.8          & 46.6          & 59.5          & 35.9          & 48.1          \\
HippoRAG (GPT-4o)             & 43.7          & 57.7          & 26.9          & 39.0          & 59.4          & 67.9          & 43.3          & 54.9          \\
IRCoT+HippoRAG (GPT3.5)*       & 45.7          & 59.2          & 21.9          & 33.3          & 47.7          & 62.7          & 38.4          & 51.7          \\
IRCoT+HippoRAG (GPT-4o)       & 49.2          & 64.0          & 31.5          & 44.2          & 68.0          & 77.2          & 49.6          & 61.8          \\
\midrule
\textbf{EviPath (w/o Golden Evid., Llama3.1-8B)} & \underline{51.8}    & \underline{65.0}    & 34.0    & 44.3          & \underline{91.1}    & \underline{93.4}    & 59.0    & 67.6    \\
\textbf{EviPath (Llama3.2-3B)}  & 48.6 & 60.7 & \underline{39.9} & \underline{48.8} & 90.4 & 92.9 & \underline{59.6} & \underline{67.4} \\
\textbf{EviPath (Llama3.1-8B)}                       & \textbf{53.8} & \textbf{66.4} & \textbf{44.3} & \textbf{54.6} & \textbf{91.3} & \textbf{93.6} & \textbf{63.1} & \textbf{71.5} \\
\bottomrule
\end{tabular}
\label{tab:pipeline_rag}
\end{table*}

\section{Discussion}\label{app:discussion}
\textbf{RQ9: Are existing state-of-the-art commercial LLMs capable of mastering complex question answering with specified RAG pipelines or workflows?} 
In order to answer this question, we re-examine two key state-of-the-art pipeline RAG methods, namely IRCoT~\citep{IRCoT} and HippoRAG~\citep{HippoRAG} with the state-of-the-art commercial LLM: GPT-4o. %Table~\ref{tab:pipeline_rag} shows the experimental results.
From the experimental results in Table~\ref{tab:pipeline_rag}, we observe that replacing the backbone model with the more advanced GPT-4o leads to some performance improvements in pipeline RAG systems. However, these gains are insufficient to surpass the RAG agent developed on synthetic data produced by EviPath. Despite benefiting from the combined strengths of IRCoT’s chain-based retrieval and HippoRAG’s memory indexing, GPT-4o still lags behind the EviPath-trained 8B RAG agent by $13.5\%$ in EM. 

Hence, we can conclude that: (i) pipeline RAG systems rely intensively on the capacity of backbone models, which often requires ultra-large LLMs like GPT3-175B, GPT3.5, and GPT-4o to deliver adequate performance. (ii) In contrast, EviPath-synthesized trajectories empower small-scale LLMs to surpass their capability ceilings, yielding significant improvements in complex question answering. 
\\

\textbf{RQ10: Are RAG agents trained by synthetic reasoning trajectories robust enough when encountering unexpected situations (e.g. incomplete retrieval and failed plans)?} In real-world scenarios, RAG agents may receive negative feedback from the external environment. Specifically, we encounter incomplete retrieval results in $978$ out of the $7,405$ questions in the HotpotQA development set. However, our RAG agents are still able to correctly answer 291 of these questions, achieving an accuracy of $29.8\%$, which demonstrates strong robustness. From examining the actual trajectory outputs, we observe that our RAG agent primarily employs two strategies to cope with incomplete or failed retrieval: (i) rephrasing the sub-question and issuing an additional search query, and (ii) making inferences or educated guesses based on its internal knowledge.

\onecolumn
\newpage
\section{Prompt Templates of Data Synthesis}
\label{app:synthesis_prompts}
\begin{longtable}{@{}p{\dimexpr\textwidth-2\tabcolsep\relax}@{}}

\caption{Data synthesis prompt template for the planner module}
\label{tab:musique_synthesis_prompt_corrected}\\
\toprule
% \endfirsthead marks the end of the header for the *first page*.
\textbf{\#\# System Prompts \#\#} \\
\midrule
\endfirsthead

% \endhead marks the end of the header for all *subsequent pages*.
\caption[]{Data synthesis prompt template for the Planner module (continued)}\\
\toprule
% \textbf{\#\# Assistant (continued) \#\#} \\
% \midrule
\endhead

% \endfoot marks the end of the footer for all pages *except the last one*.
\midrule
\multicolumn{1}{r}{\textit{Continued on next page...}} \\
\endfoot

% \endlastfoot marks the end of the footer for the *very last page*.
\bottomrule
\endlastfoot

% --- Main Table Content Begins Here ---
Your task is to provide a reasoning path to answer a complex question. \\
You will be given a question, the answer of the question, and supporting evidence to obtain the answer. \\
You need to:\\
1. Make a plan about how to answer this question by decompose it into multiple steps. \\
2. Solve the problem in a step-by-step manner.\\
Put your thoughts between \texttt{<think>} and \texttt{</think>} tags. \\
You can use the \texttt{Search(entity, "question")} function to get information.\\
Extract candidate entity from the intermediate subquestions with \texttt{[X]}. \\
Label the final answer with \texttt{<answer> *** </answer>}. \\
\midrule
\textbf{\#\# User \#\#} \\
Question: Which film has the director born earlier, Ek Paheli or When Father Was Away On Business?\\
Answer: Ek Paheli\\
Supporting evidences: \\
(Ek Paheli, director, Naresh Kumar)\\
(When Father Was Away on Business, director, Emir Kusturica)\\
(Naresh Kumar (tennis), date of birth, 22 December 1928) \\
(Emir Kusturica, date of birth, 24 November 1954)\\
\\
Question: Which film has the director born earlier, Ek Paheli or When Father Was Away On Business? Candidate: [0] Ek Paheli, [1] When Father Was Away On Business \\
Answer: Ek Paheli \\
\midrule
\textbf{\#\# Assistant \#\#} \\
\texttt{<think>}\\
To solve this problem, I need to:\\
1. Find the director of film Ek Paheli.\\
2. Find the director of film When Father Was Away On Business.\\
3. Figure out the date of birth of the director of Ek Paheli.\\
4. Figure out the date of birth of the director of When Father Was Away On Business.\\
5. Compare their dates of birth to determine which director was born earlier.\\
\texttt{</think>}\\
\texttt{<action>}\\
Search([0], "Who is the director of Ek Paheli?")\\
Search([1], "Who is the director of When Father Was Away On Business?")\\
\texttt{</action>}\\
\midrule
\textbf{\#\# User \#\#}\\
Observation: \\
The director of Ek Paheli is Naresh Kumar. Candidate: [2] Naresh Kumar\\
The director of When Father Was Away On Business is Emir Kusturica. Candidate: [3] Emir Kusturica\\
\midrule
\textbf{\#\# Assistant \#\#} \\
\texttt{<think>}\\
Now that I have the directors of both films, I need to find their dates of birth.\\
\texttt{</think>}\\
\texttt{<action>}\\
Search([2], "What is the date of birth of Naresh Kumar?")\\
Search([3], "What is the date of birth of Emir Kusturica?")\\
\texttt{</action>}\\
\midrule
\textbf{\#\# User \#\#}\\
Observation: \\
Naresh Kumar was born on 22 December 1928. \\
Emir Kusturica was born on 24 November 1954. \\
\midrule
\textbf{\#\# Assistant \#\#} \\
\texttt{<think>} \\
Now that I have the birth dates of the two directors. Since Naresh Kumar was born earlier than Emir Kusturica, the director of Ek Paheli was born earlier. \\
\texttt{</think}\\
\texttt{<answer> Ek Paheli </answer>} \\
\end{longtable}

\vspace{3em}
\begin{table}[h]
\caption{Data synthesis prompt template for the Executor module}
\centering
\resizebox{\textwidth}{!}{
\begin{tabular}{p{\textwidth}}
\toprule
\textbf{\#\# System Prompts \#\#} \\
Your task is to provide reasoning path to answer a question based on some supporting evidence. \\
Note that there are some disturbance terms within the supporting evidence.\\
You will be given a question, the answer of the question, and the index of golden supporting evidence to obtain the answer.\\
You need to: \\
1. Identify the requested information and summarize the information provided in the golden evidence. \\
Please put your thoughts between \texttt{<think>} and \texttt{</think>} tags. Please do NOT consider other supporting evidence other than the golden evidence. \\
2. Identify the index of the ground truth supporting evidence. Put the index between \texttt{<select>} and \texttt{</select>} tags.\\
3. Summarize the answer with one sentence. Put the answer between \texttt{<answer>} and \texttt{</answer>} tags. \\
\midrule
\textbf{\#\# User \#\#} \\
Question: Who is the mother of Xawery Żuławski?\\
Supporting evidence:\\{}
[0] (Xawery Żuławski, mother, Małgorzata Braunek)\\{}
[1] (Xawery Żuławski, father, Andrzej Żuławski)\\{}
[2] (Xawery Żuławski, family, Q63532193)\\{}
[3] (Xawery Żuławski, family name, Q56541485)\\{}
[4] (Xawery Żuławski, spouse, Maria Strzelecka)\\{}
[5] (Xawery Żuławski, date of birth, 1971-12-22T00:00:00Z)\\{}
[6] (Xawery Żuławski, sibling, Vincent Zulawski)\\{}
[7] (Xawery Żuławski, place of birth, Warsaw)\\{}
[8] (Andrzej Żuławski, child, Xawery Żuławski)\\{}
[9] (Małgorzata Braunek, child, Xawery Żuławski)\\{}
Answer: Małgorzata Braunek\\
Golden evidence: [0] (Xawery Żuławski, mother, Małgorzata Braunek)\\

\midrule
\textbf{\#\# Assistant \#\#} \\
\texttt{<think>} The question asks me to find the mother of Xawery Żuławski. [0] says Xawery Żuławski's mother Małgorzata Braunek. \texttt{</think>} \\
\texttt{<select>} [0] \texttt{</select>} \\
\texttt{<answer>} The mother of Xawery Żuławski is Małgorzata Braunek. \texttt{</answer>} \\
\bottomrule
\end{tabular}
}
\label{tab:executor_synthesis}
\end{table}

\newpage

\section{Prompt Templates of RAG Agent Training}
\label{app:training_prompts}
\begin{table}[h!]
\caption{Planner Prompt Template  for text-based multi-hop question answering}
\centering
\resizebox{\textwidth}{!}{
\begin{tabular}{p{\textwidth}}
\toprule
\textbf{\#\# System Prompts \#\#} \\
Your task is to provide a reasoning path to answer a complex question. \\
You will be given a question and the answer of the question.\\
You need to:\\
1. Make a plan about how to answer this question by decompose it into multiple steps. \\
2. Solve the problem in a step-by-step manner.\\
Put your thoughts between \texttt{<think>} and \texttt{</think>} tags. \\
You can use the \texttt{Search("question")} function to get information from supporting evidence.\\
Please note the the content within the Search function should be a question (interrogative sentence).\\
When issuing a search query, please refrain from using pronouns like ``this'', ``these'', etc., and use the full entity names instead.\\
Label the final answer within \texttt{<answer>} *** \texttt{</answer>}. \\
\midrule
\textbf{\#\# User \#\#} \\
Question: When was the institute that owned The Collegian founded? \\
\midrule
\textbf{\#\# Assistant \#\#} \\
\texttt{<think>} \\
To solve this problem, I need to: \\
1. Find the institute that owned The Collegian. \\
2. Figure out when the institute that owns The Collegian was founded. \\
Now I need to search the information about The Collegian first. \\
\texttt{</think>} \\
\texttt{<action> Search("Which institute owns The Collegian?") </action>} \\
\midrule
\textbf{\#\# User \#\#} \\
Obs: The Collegian is owned by Houston Baptist University. \\
\midrule
\textbf{\#\# Assistant \#\#} \\
\texttt{<think>} \\
Now that I have the owner of The Collegian, I need to figure out when Houston Baptist University was founded. \\
\texttt{</think>} \\
\texttt{<action> Search("When was Houston Baptist University founded?") </action>} \\
\midrule
\textbf{\#\# User \#\#} \\
Obs: Houston Baptist University was founded in 1960. \\
\midrule
\textbf{\#\# Assistant \#\#} \\
\texttt{<think>} \\
OK. Now I have all the necessary information to answer the question. The question asking when the institute that owned The Collegian was founded. \\
\texttt{</think>} \\
\texttt{<answer> 1960 </answer>} \\
\bottomrule
\end{tabular}
}
\label{planner_musique}
\end{table}

\newpage
\begin{table}[h!]
\caption{Executor Prompt Template  for Text-based Multi-hop Question Answering}
\centering
\resizebox{\textwidth}{!}{
\begin{tabular}{p{\textwidth}}
\toprule
\textbf{\#\# System Prompts \#\#} \\
Your task is to answer a question based on some supporting evidence. \\
Note that there are some disturbance terms within the supporting evidence. \\
You will be given a question and a list of supporting evidence. \\
You need to: \\
1. Identify the requested information from the question. Review all supporting evidence, summarize the information provided in evidences that support answering the question. Please put your thoughts between \texttt{<think>} and \texttt{</think>} tags. \\
2. Identify the index of the golden supporting evidence. Put the index between \texttt{<select>} and \texttt{</select>} tags. If multiple supporting evidence contain the answer, select all of them. If there are no evidence matches, respond with "No relevant information found." and do not output any other contents. \\
3. Summarize the answer with one complete declarative sentence. Put the answer between \texttt{<answer>} and \texttt{</answer>} tags. \\
Please use the following template: \\
\texttt{<think> ... </think>} \\
\texttt{<select> [X] </select>} \\
\texttt{<answer> The complete answer sentence. </answer>} \\
\midrule
\textbf{\#\# User \#\#} \\
Question: When was magazine LaIsha founded? \\
Supporting evidences: \\{}
[0] LaIsha: La\'Isha (also known as "For the Woman") is an Israeli magazine for girls and boys. \\{}
[1] LaIsha: It has been published on weekly basis since 1947, and is owned by Yedioth Ahronoth media group. \\{}
[2] Glamour (magazine): Glamour is a women's magazine published by Condé Nast Publications. \\{}
[3] Glamour (magazine): Founded in 1939 and first published in April 1939 in the United States, it was originally called "Glamour of Hollywood" \\{}
[4] Salt to the Sea: Salt to the Sea is a 2016 historical fiction young adult novel by Ruta Sepetys. \\{}
[5] Salt to the Sea: It tells the story of four individuals in World War II who make their way to the ill-fated MV "Wilhelm Gustloff". \\{}
[6] Salt to the Sea: The story also touches on the disappearance of The Amber Room, a work of art stolen by the Nazis that has never been recovered.\\{}
[7] MV Wilhelm Gustloff: MV "Wilhelm Gustloff" was a German military transport ship which was sunk on 30 January 1945 by in the Baltic Sea while evacuating German civilians, Nazi officials and military personnel from Gdynia (Gotenhafen) as the Red Army advanced.\\{}
[8] MV Wilhelm Gustloff: By one estimate, 9,400 people died, which makes it the largest loss of life in a single ship sinking in history\\
\midrule
\textbf{\#\# Assistant \#\#} \\
\texttt{<think>} The question asks me to determine when the magazine LaIsha was founded. [1] says the magazine LaIsha was founded in 1947. \texttt{</think>} \\
\texttt{<select>} [1] \texttt{</select>} \\
\texttt{<answer>} LaIsha was founded in 1947. \texttt{</answer>} \\
\bottomrule
\end{tabular}
}
\label{worker_mrc}
\end{table}

\end{document}